\documentclass{article}
\usepackage{amsmath,epsfig,bm, amssymb,graphicx,algorithm,algorithmic,color}
\newtheorem{defi}{Definition}
\newtheorem{prop}[defi]{Proposition}

\newcommand{\proofend}{\hfill$\Box$\vspace{2mm}}

\newcommand{\argmin}{\mathop{\mathrm{argmin\,}}}

\newcommand{\mathbbR}{\mathbb{R}}

\newcommand{\boldone}{{\boldsymbol{1}}}

\newcommand{\boldB}{{\boldsymbol{B}}}

\newcommand{\boldP}{{\boldsymbol{P}}}

\newcommand{\boldX}{{\boldsymbol{X}}}

\newcommand{\bolda}{{\boldsymbol{a}}}
\newcommand{\boldb}{{\boldsymbol{b}}}

\newcommand{\boldw}{{\boldsymbol{w}}}
\newcommand{\boldx}{{\boldsymbol{x}}}

\newcommand{\boldz}{{\boldsymbol{z}}}

\newcommand{\boldPi}{{\boldsymbol{\Pi}}}

\newcommand{\calD}{{\mathcal{D}}}

\newcommand{\calM}{{\mathcal{M}}}

\newcommand{\calS}{{\mathcal{S}}}
\newcommand{\calT}{{\mathcal{T}}}
\newcommand{\calU}{{\mathcal{U}}}

\newcommand{\calX}{{\mathcal{X}}}

\newcommand{\Nsample}{N}

\usepackage[noorphans]{quoting}

\usepackage[square,sort,comma,numbers]{natbib}

\usepackage{setspace}
\usepackage{wrapfig}

\usepackage{subcaption}

\usepackage{url,multirow}

\advance\oddsidemargin-1in
\date{\today}
\title{Approximating 1-Wasserstein Distance with Trees} 
 
\author{%
   Makoto Yamada$^{1,2}$, Yuki Takezawa$^{1,2}$, Ryoma Sato$^{1,2}$, Han Bao$^{1}$\\
   Zornitsa Kozareva$^3$, Sujith Ravi$^4$\\
   $^1$Kyoto University, $^2$RIKEN AIP, $^3$Meta AI, $^4$SliceX AI\\
}

\begin{document}
\maketitle

\begin{abstract}
Wasserstein distance, which measures the discrepancy between distributions, shows efficacy in various types of natural language processing (NLP) and computer vision (CV) applications. One of the challenges in estimating Wasserstein distance is that it is computationally expensive and does not scale well for many distribution comparison tasks. In this paper, we aim to approximate the 1-Wasserstein distance by the tree-Wasserstein distance (TWD), where TWD is a 1-Wasserstein distance with tree-based embedding and can be computed in linear time with respect to the number of nodes on a tree. More specifically, we propose a simple yet efficient L1-regularized approach to learning the weights of the edges in a tree. To this end, we first show that the 1-Wasserstein approximation problem can be formulated as a distance approximation problem using the shortest path distance on a tree. We then show that the shortest path distance can be represented by a linear model and can be formulated as a Lasso-based regression problem. Owing to the convex formulation, we can obtain a globally optimal solution efficiently. Moreover, we propose a tree-sliced variant of these methods. Through experiments, we demonstrated that the weighted TWD can accurately approximate the original 1-Wasserstein distance.
\end{abstract}

\section{Introduction}
Wasserstein distance, which is an optimal transport (OT)-based distance, measures the discrepancy between two distributions, and is widely used in natural language processing (NLP) and computer vision (CV) applications. For example, measuring the similarity between documents is a fundamental natural language processing task, and can be used for semantic textual similarity (STS) tasks \citep{yokoi2020word}. For CV tasks, because it is possible to obtain matching between samples using OT, it is used to determine the correspondence between two sets of local features \citep{sarlin2020superglue}.

One of the widely used application of the Wasserstein distance is Word Mover's Distance (WMD) \citep{kusner2015word}, which measures the similarity between documents by solving the optimal transport problem. Recently, \citet{yokoi2020word} proposed a more effective similarity measure based on the Word Rotator's Distance (WRD). The WMD and WRD are distance-based optimal transport, which can be estimated by solving linear programming with $O(n^3)$ order of computation, where $n$ is the number of words in a document. Thus, using WMD and WRD for a large number of documents is challenging. 

One of the well-known speedup techniques for OT problems is the use of the Sinkhorn algorithm \citep{cuturi2013sinkhorn}, which solves the entropic regularized optimal transport problem. Using the Sinkhorn algorithm, we can estimate WMD and WRD with $O(n^2)$ computational cost. However, this is slow for most NLP and CV tasks. Instead of using linear programming and the Sinkhorn algorithm, we can speed up the computation by projecting word vectors into 1D space and solving the OT problem. This approach is known as the sliced Wasserstein distance \citep{kolouri2016sliced}, and its computational complexity is $O(n \log n)$. However, performance is highly dependent on the projection matrix. 

Tree-Wasserstein distance (TWD) \citep{evans2012phylogenetic,le2019tree} computes the optimal transport on a tree, and it can be computed in $O(N)$, where $N$ is the number of nodes in a tree. Specifically, we first embedded the word vectors into a tree and computed the optimal transport based on a tree metric. Because we can obtain an analytical solution to the OT problem on a tree, it can be solved efficiently. Moreover, TWD includes sliced Wasserstein as a special case, and it empirically outperforms sliced Wasserstein \citep{le2019tree}. The performance of TWD is highly dependent on tree construction. One of the widely used tree construction methods is based on the QuadTree \citep{indyk2003fast}. Another approach involves using a clustering-based tree algorithm \citep{le2019tree}. Recently, \citet{takezawa2021supervised} proposed a continuous tree-construction approach. Although tree-based methods improve the computational complexity by a large margin, existing tree-based methods do not approximate the vanilla Wasserstein distance satisfactorily.

In this paper, we aimed to accurately approximate the 1-Wasserstein distance using TWD. Specifically, we propose a simple yet efficient sparse learning-based approach to train the edge weights of a tree. To this end, we first demonstrate that the 1-Wasserstein approximation problem can be formulated as a distance approximation problem using the shortest path distance on a tree. Then, we show that the shortest path distance can be represented by a linear model when we fix the tree structure. The weight estimation problem is then formulated as a non-negative Lasso-based optimization problem. Owing to the convex formulation, we can obtain a globally optimal solution efficiently. Moreover, we propose a tree-sliced variant of the proposed method. One of the key advantages of our formulation is that the optimization can be efficiently solved using an off-the-shelf solver. Through experiments, we evaluate the proposed algorithm for Twitter, BBCSport, and Amazon datasets and show that the 1-Wasserstein distance computed by linear programming can be accurately approximated by TWD, whereas QuadTree \citep{indyk2003fast} and ClusterTree \citep{le2019tree} have high approximation errors.

\vspace{.05in}
\noindent {\bf Contribution:} The contributions of our paper are summarized below:
\begin{itemize}
    \item A Lasso-based weight estimation is proposed for TWD.    \item An estimation of the weights for tree-sliced Wasserstein distance is proposed.    \item We empirically demonstrate that the proposed method is capable of approximating the 1-Wasserstein distance accurately.
\end{itemize}

\section{Related Work}

Measuring the distance between sets is an important research topic in machine learning, computer vision, and natural language processing. Among the early works, Earth Mover's Distance (EMD) is an optimal transport-based distance, equivalent to Wasserstein distance if we employ distance to compute the cost function and can be computed using linear programming. EMD has been particularly studied in the computer vision community because it can obtain matching between samples. For example, \citet{sarlin2020superglue} proposed the SuperGlue method, which determines the correspondence between two sets of local features using optimal transport. \citet{liu2020semantic} uses the optimal transport method for semantic correspondence. In NLP tasks, \citet{kusner2015word} proposed Word Mover's Distance (WMD), the first work to use Wasserstein distance for textual similarity tasks, and it is widely used in NLP tasks, including text generation evaluation \citep{zhao2019moverscore}. Recently, \citet{sato2021re} re-evaluated WMD in various experimental setups and summarized the pros and cons of WMD. Another promising approach is based on the Word Rotater's Distance (WRD) \citep{yokoi2020word}, which normalizes the word vectors and solves WMD with adjusted probability mass, and it has been reported that the performance significantly improves over WMD. However, these approaches employ EMD, and the computational cost is very high; therefore, they cannot be used for large-scale distribution comparison tasks. 

To speed up EMD and Wasserstein distance computation, \citet{cuturi2013sinkhorn} proposed the Sinkhorn algorithm, which solves the entropic regularized optimization problem and computes Wasserstein distance in quadratic order (i.e., $O(n^2)$).  \citep{kusner2015word} proposed a relaxed Wasserstein Mover's distance (RWMD), which only considers the some to one constraint. Because RWMD computation is simple, it can efficiently compute the distance. \citep{atasu2019linear} proposed the approximate constrained transfer method, which iteratively adds constraints to RWMD and improves the performance of RWMD while maintaining the computational speed.

Another approach is based on the sliced Wasserstein distance (SWD) \citep{kolouri2016sliced}, which solves the optimal transport problem in a projected one-dimensional subspace. Because it is known that one-dimensional optimal transport can be solved using sorting, SWD can be computed in $O(n \log n)$. Several extensions of SWD have been proposed. The generalized sliced-Wasserstein distance extends the 1D sliced Wasserstein distance for multidimensional cases \citep{kolouri2019generalized}.
The max-sliced Wasserstein distance determines the 1D subspace that maximizes the discrepancy between two distributions and then computes the optimal transport in the subspace \citep{deshpande2019max}. The subspace robust Wasserstein distance (SRWD) is a general extension of the max-sliced Wasserstein distance for multi-dimensional subspaces \citep{paty2019subspace}.

Tree-Wasserstein distance (TWD) \citep{evans2012phylogenetic,le2019tree} uses tree-based embedding, whereas SWD uses one-dimensional embedding. Because a chain of a tree can represent one-dimensional embedding, TWD includes SWD as a special case. \citet{le2019tree} reported that the TWD can empirically outperform SWD. TWD is also studied in the theoretical computer science community and can be computed using the QuadTree algorithm \citep{indyk2003fast}. The extension of TWD has also been studied, including unbalanced TWD \citep{sato2020fast,le2021entropy}, supervised Wasserstein training \citep{takezawa2021supervised}, and tree barycenter \citep{takezawa2021fixed}.  These approaches mainly focus on approximating 1-Wasserstein with tree construction, and the weight of the edges are set by a constant number. Recently, \citet{backurs2020scalable} proposed a flowtree tree that combines the QuadTree method and the cost matrix computed from vectors. They then showed that the flowtree outperformed QuadTree-based approaches. Moreover, they theoretically guarantee that the QuadTree and flowtree methods can approximate the nearest neighbor using the 1-Wasserstein distance.  \citep{dey2022approximating} proposed an L1-embedding approach for approximating 1-Wasserstein distance for the persistence diagram. However, there are no learning-based approaches for approximating 1-Wasserstein for general setup. Thus, we focus on estimating the weight parameter of TWD from the data to approximate 1-Wasserstein distance.

\section{Preliminary}
In this section, we introduce the optimal transport, Wasserstein distances, and tree-Wasserstein distances.
\subsection{Optimal transport}
We computed the distance between two datasets $\{(\boldx_i, a_i)\}_{i = 1}^n$ and $ \{(\boldx'_i, b_i)\}_{j = 1}^{n'}$, where $\boldx \in \calX \subset \mathbbR^d$,  $\boldx' \in \calX' \subset \mathbbR^d$, and $\sum_{i = 1}^n a_i = 1$ and $\sum_{j = 1}^{n'} b_j = 1$ are the probability masses. For example, in the text similarity setup, $\boldx$ and $\boldx'$ are precomputed word-embedding vectors, and $a_i$ is the frequency of the word $\boldx_i$ in a document. Let $\calD = \{\boldx_i\}_{i = 1}^{\Nsample'}$ denote the entire set of vectors, where $\Nsample'$ denotes the number of vectors. The goal of this paper is to measure the similarity between two datasets $\{(\boldx_i, a_i)\}_{i = 1}^n$ and $ \{(\boldx'_i, b_i)\}_{j = 1}^{n'}$.

 In the Kantorovich relaxation of OT, admissible couplings are defined by the set of transport plans between two discrete measures, $\mu = \sum_{i=1}^{n} a_i \delta_{\boldx_i}$ and $\nu = \sum_{j=1}^{n'} b_j \delta_{\boldx'_j}$.
\begin{align*}
U(\mu,\nu) = \{ \boldPi \in \mathbbR_{+}^{n \times n'} : \boldPi \boldone_{n'} = \bolda, \boldPi^\top \boldone_n = \boldb \},
\end{align*}
where $\delta_{\boldx_i}$ is the Dirac at position $\boldx_i$, $\boldone_n$ is the $n$-dimensional vector whose elements are ones, and $\bolda = (a_1, a_2, \ldots, a_n)^\top \in \mathbbR^{n}_+$ and $\boldb = (b_1, b_2, \ldots, b_{n'})^\top \in \mathbbR^{n'}_+$ are the probability masses of distributions $\mu$ and $\nu$, respectively.

Then, the OT problem between two discrete measures $\mu$ and $\nu$ is given as
\begin{align}
\label{eq:ot}
    \min_{\boldPi \in U(\mu,\nu)}&\hspace{.3cm} \sum_{i=1}^{n}\sum_{j=1}^{n'} \pi_{ij}  c(\boldx_i, \boldx'_j),
\end{align}
where $\pi_{ij}$ is the $i,j$-th element of $\boldPi$, and $c(\boldx,\boldx')$ is a cost function (e.g., $c(\boldx,\boldx') = \|\boldx - \boldx'\|_2^2$). Eq. \eqref{eq:ot} is equivalent to the Earth Mover's Distance and it can be solved by using linear programming with $O(n^3)$ computation  $(n = n')$. To speed up the optimal transport computation, \citet{cuturi2013sinkhorn} proposed an entropic regularized optimal transport problem:
\begin{align*}
\min_{\boldPi \in U(\mu,\nu)}&\hspace{.3cm} \sum_{i=1}^{n}\sum_{j=1}^{n'} \pi_{ij} c(\boldx_i, \boldx'_j) - \epsilon H(\boldPi),
\end{align*}
where $\epsilon \geq 0$ is the regularization parameter and $H(\boldPi) = -\sum_{i=1}^n \sum_{j = 1}^{n'} \pi_{ij}(\log(\pi_{ij}) - 1)$ is the entropic regularization. The entropic regularized OT is equivalent to EMD when $\epsilon = 0$. The entropic regularization problem can be efficiently solved using the Sinkhorn algorithm with $O(nn')$ cost. Moreover, the Sinkhorn algorithm can be efficiently implemented on GPU, leading to a super-efficient matrix computation. Hence, the Sinkhorn algorithm is one of the most widely used OT solvers in the ML field. 

\subsection{Wasserstein Distance} If we use distance function $d(\boldx,\boldx')$ for the cost $c(\boldx,\boldx')$ and $p \geq 1$, the $p$-Wasserstein distance of two discrete measures between two probability measures $\mu$ and $\nu$ is defined as
\begin{align*}
    W_p(\mu, \nu) = \left(\inf_{\pi \in \calU(\mu, \nu)} \int_{\calX \times \calX'} d(\boldx, \boldx') \textnormal{d}\pi(\boldx,\boldx')^p\right)^{1/p},
\end{align*}
where $\calU$ denotes a set of joint probability distributions \citep{peyre2019computational} (Remark 2.13):
\begin{align*}
    \calU(\mu,\nu)=\{ \pi \in \calM_+^1(\calX \times \calX') : P_{\calX\sharp}\pi = \mu, P_{\calX'\sharp}\pi = \nu \},
\end{align*}
and $\calM_+^1(\calX)$ is the set of probability measures, $P_{\calX\sharp}$ and $P_{\calX'\sharp}$ are push-forward operators (see Remark 2.2 of \citep{peyre2019computational}).

The $p$-Wasserstein distance of the two discrete measures $\mu = \sum_{i=1}^{n} a_i \delta_{\boldx_i}$ and $\nu = \sum_{j=1}^{m} b_j \delta_{\boldx'_j}$ is defined as
\begin{align*}
    W_p(\mu, \nu) = \left(\min_{\boldPi \in U(\mu,\nu)} \sum_{i=1}^{n}\sum_{j=1}^{n'} \pi_{ij}  d(\boldx_i, \boldx'_j)^p\right)^{1/p}.
\end{align*}
From the definition of the $p$-Wasserstein distance, the 1-Wasserstein distance of two discrete measures is given as:
\begin{align*}
    W_1(\mu, \nu) = \min_{\boldPi \in U(\mu,\nu)} \sum_{i=1}^{n}\sum_{j=1}^{n'} \pi_{ij}  d(\boldx_i, \boldx'_j),
\end{align*}
which is equivalent to Eq. \eqref{eq:ot} when $c(\boldx,\boldx') = d(\boldx,\boldx')$. For the distance function, the Euclidean distance $d(\boldx, \boldx') = \|\boldx - \boldx'\|_2$ and Manhattan distance $d(\boldx,\boldx') = \|\boldx - \boldx'\|_1$ are common choices. Note that \citet{cuturi2013sinkhorn} proposed the entropic regularized Wasserstein distance, although the entropic regularized formulation does not admit the triangle inequality, and thus, it is not a distance. Sinkhorn divergence was proposed to satisfy the metric axioms of the entropic regularized formulation \citep{genevay2018learning}.


\subsection{Tree-Wasserstein distance (TWD)}
Although the Sinkhorn algorithm has quadratic time complexity, it is still prohibitive when dealing with large vocabulary sets. TWD has been gathering attention owing to its light computation, and is thus used to approximate the 1-Wasserstein distance \citep{indyk2003fast,evans2012phylogenetic,le2019tree}.

We define $\calT= (V,E)$, where $V$ and $E$ are the sets of nodes and edges, respectively. In particular, for a given tree $\mathcal{T}$, EMD with a tree metric $d_\mathcal{T}(x, x')$ is called TWD. TWD admits the following closed-form expression:
\begin{align*}
    W_\calT(\mu,\nu) = \sum_{e \in E} w_e |\mu(\Gamma(v_e)) - \nu(\Gamma(v_e))|,
\end{align*}
where $e$ is an edge index, $w_e \in \mathbbR_+$ is the edge weight of edge $e$, $v_e$ is the $e$-th node index, and $\mu(\Gamma(v_e))$ is the total mass of the subtree with root $v_e$.  The restriction to a tree metric is advantageous in terms of computational complexity, TWD can be computed as $O(\Nsample)$, where $\Nsample$ is the number of nodes of a tree. It has been reported that TWD compares favorably with the vanilla Wasserstein distance and is computationally efficient. Note that the sliced Wasserstein distance can be regarded as a special case of TWD \citep{takezawa2021fixed}.

\begin{figure}[t]
    \centering
    \begin{subfigure}[t]{0.45\textwidth}
        \centering
        \includegraphics[width=0.99\textwidth]{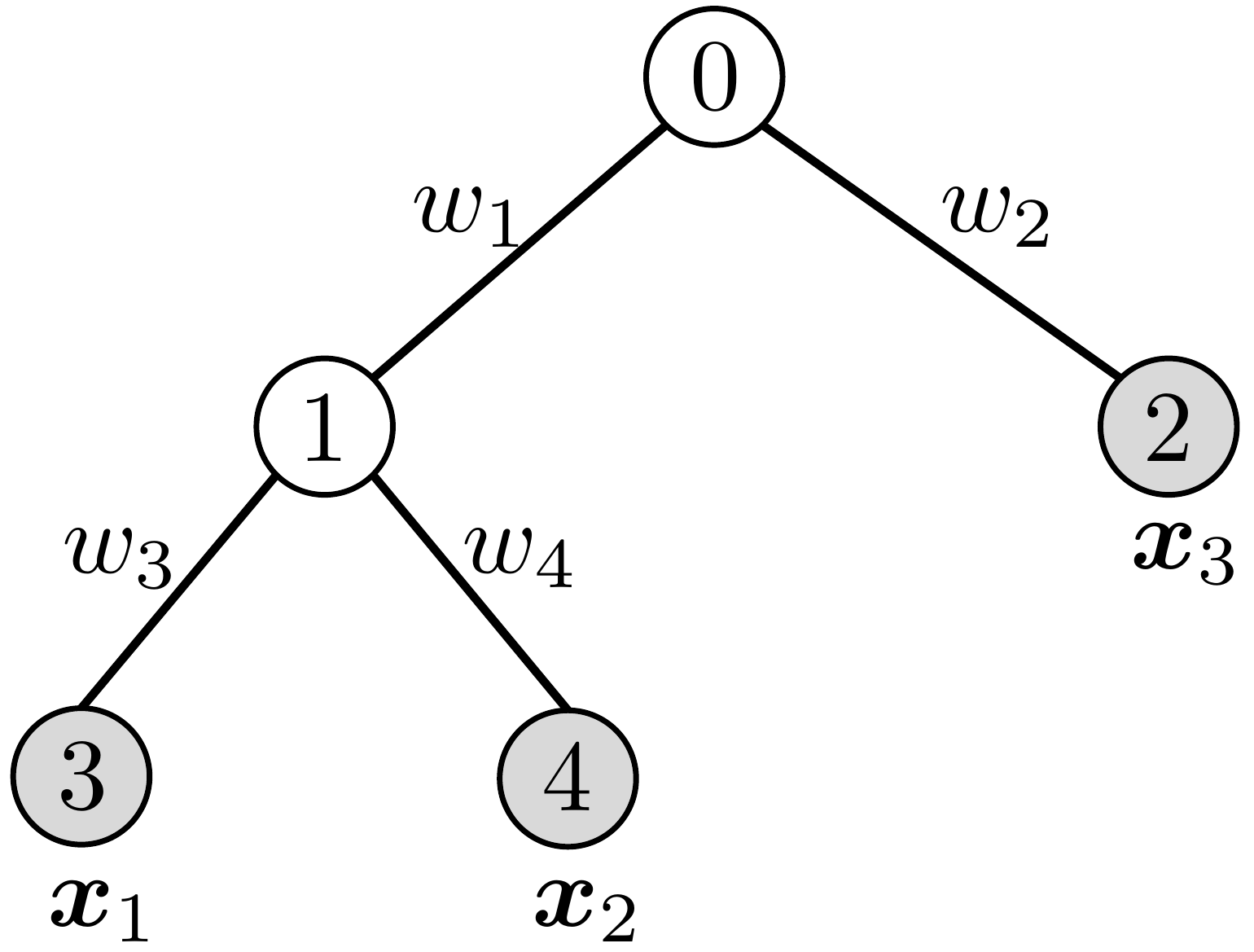}
        \caption{An illustration of tree embedding. \label{fig:tree}}
    \end{subfigure}\quad
    \begin{subfigure}[t]{0.45\textwidth}
        \centering
        \includegraphics[width=0.99\textwidth]{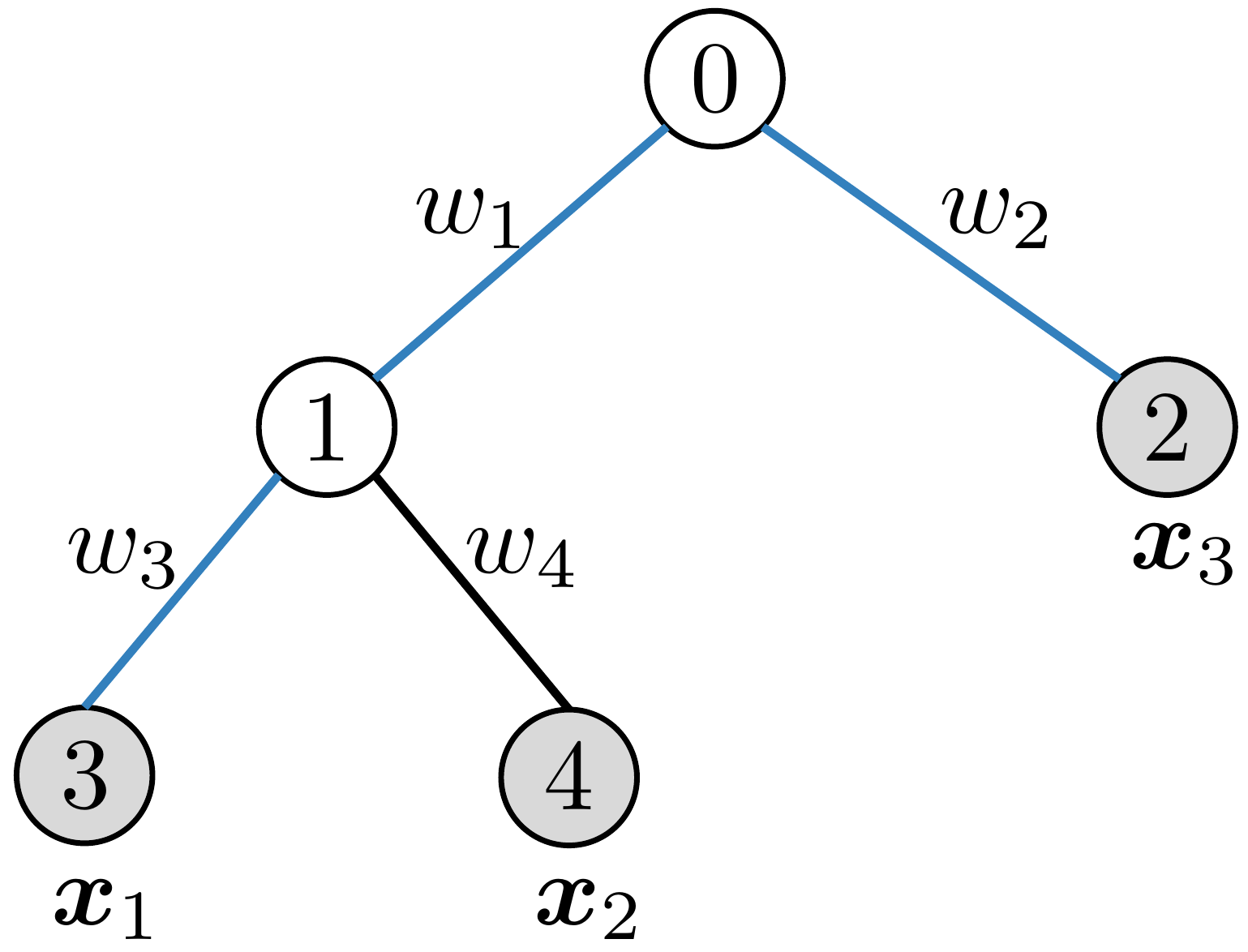}
        \caption{Shortest path between $\boldx_1$ and $\boldx_3$. \label{fig:shortest_path}}
    \end{subfigure}\quad
    \caption{An illustration of tree embedding and the shortest path between $\boldx_1$ and $\boldx_3$ of the tree. All the input vectors are associated with the leaf nodes. The $\boldb_1$, $\boldb_2$, and $\boldb_3$ of the tree are given as $\boldb_1 = (1, 1, 0, 1, 0)^\top$, $\boldb_2 = (1, 1, 0, 0, 1)^\top$, and $\boldb_3 = (1, 0, 1, 0, 0)^\top$, respectively.   In this case, $d_\calT(\boldx_1,\boldx_3) = w_3 + w_1 + w_2$.}
    \vspace{-.15in}
\end{figure}

TWD can be represented in matrix form as \citep{takezawa2021supervised}
\begin{align*}
    W_{\calT}(\mu, \nu) = \left\| \text{diag}(\boldw)\boldB(\bolda - \boldb) \right\|_1,
\end{align*}
where $\boldw \in \mathbbR_+^\Nsample$ is the weight vector, $\boldB \in \{0,1\}^{\Nsample \times \Nsample_{\text{leaf}}}$ is a tree parameter, $[\boldB]_{i,j} = 1$ if the node $i$ is the parents node of the leaf node $j$ and zero otherwise, $\Nsample$ is the total number of nodes of a tree, $\Nsample_{\text{leaf}}$ is the number of leaf nodes, and  $\text{diag}(\boldw) \in \mathbbR_+^{\Nsample \times \Nsample}$ is a diagonal matrix whose diagonal elements are $\boldw$. Figure \ref{fig:tree} shows an example of a tree embedding. In this tree, the total number of nodes is $\Nsample = 5$ and the number of leaf nodes is $\Nsample_{\text{leaf}} = 3$. It is noteworthy that $w_0$ is the weight of the root node. Because $\bolda$ and $\boldb$ are probability vectors (i.e., $\bolda^\top \boldone = 1$), and the elements of the first row of $\boldB$ are all 1, the weight of the root node is ignored in $W_{\calT}(\mu, \nu)$. 

To compute TWD, the choice of tree structure and weight is crucial for approximating the Wasserstein distance. \citet{indyk2003fast} proposed the QuadTree algorithm, which constructs a tree by recursively splitting a space to four regions (quad) and setting the weight parameter as $2^{-\ell(e)}$ ($\ell(e)$ is the depth of an edge $e$). For QuadTree, TWD is defined as
\begin{align*}
    W_\calT(\mu,\nu) = \sum_{e \in E} 2^{-\ell(e)} |\mu(\Gamma(v_e)) - \nu(\Gamma(v_e))|.
\end{align*}
\citet{le2019tree} proposed a clustering-based tree construction. More recently, \citet{takezawa2021supervised} proposed training a tree using continuous optimization in a supervised learning setup. 

Previous research related to TWD has focused more on constructing a tree. However, learning edge weights in TWD has not been well examined. Thus, in this paper, we propose a weight estimation procedure for TWD to approximate Wasserstein distance.

\section{Proposed method}
In this section, we propose a tree weight estimation method using a non-negative Lasso to accurately approximate the 1-Wasserstein distance.
\subsection{Tree weight estimation using Lasso}
 Because TWD is a 1-Wasserstein distance with a tree metric, it is natural to estimate the weight that can approximate the 1-Wasserstein distance with an arbitrary distance $d(\boldx,\boldx')$. Thus, we propose estimating the weight $\boldw$ by minimizing the error between the 1-Wasserstein distance with distance $d(\boldx,\boldx')$ and TWD. First, we show that TWD can approximate the vanilla Wasserstein distance arbitrarily well with some trees.
 
\begin{prop} \label{prop:expressive}
For any measure $\mu = \sum_{i=1}^{n} a_i \delta_{\boldx_i}$ and $\nu = \sum_{j=1}^{n'} b_j \delta_{\boldx'_j}$ and the distance function $d\colon \mathcal{X} \times \mathcal{X} \to \mathbb{R}$, there exists a tree $\mathcal{T}$ such that $W_1(\mu, \nu) = W_\mathcal{T}(\mu, \nu)$ and $\boldPi^*(\mu, \nu) = \boldPi^*_\mathcal{T}(\mu, \nu)$ holds, where $\boldPi^*(\mu, \nu)$ is an optimal solution to Eq. \eqref{eq:ot} with cost $c = d$, and $\boldPi^*_\mathcal{T}(\mu, \nu)$ is an optimal solution with the cost function being the shortest path distance $d_\mathcal{T}$ on tree $\mathcal{T}$.
\end{prop}

The proof is provided in section \ref{sec:proof-expressive}. This proposition indicates that TWD has sufficient expressive power to model the $1$-Wasserstein distance.

\begin{prop} \citep{le2019tree}
\begin{align*}
    W_{\calT}(\mu, \nu) = \inf_{\pi \in \calU(\mu, \nu)} \int_{\calX \times \calX'} d_\calT(\boldx, \boldx') \textnormal{d}\pi(\boldx,\boldx'),
\end{align*}
where $d_\calT(\boldx,\boldx')$ is the length of the (unique) path between $\boldx$ and $\boldx'$ on a tree (i.e., the shortest path on the tree) and $\calU$ is a set of joint probability distributions. 
\end{prop}
This proposition indicates that, if $d(\boldx,\boldx') = d_\calT(\boldx,\boldx')$ and $W_1(\mu,\nu) = W_\calT(\mu,\nu)$. Thus, we aim to fit $d(\boldx,\boldx')$ using $d_\calT(\boldx,\boldx')$, where the tree structure and edge weights are potential learning parameters. Figure \ref{fig:shortest_path} shows the shortest distance between $\boldx_1$ and $\boldx_3$. In this case, $d_\calT(\boldx_1,\boldx_3) = \sum_{k \in \textnormal{Path}(\boldx_1,\boldx_3)} w_k = w_3 + w_1 + w_2$; Thus, a closed-form expression for the shortest path distance can be obtained as follows.

\begin{prop}
\label{prop:shortest_distance}
We denote by $\boldB \in \{0,1\}^{\Nsample \times \Nsample_{\text{leaf}}} = [\boldb_1, \boldb_2, \ldots, \boldb_{\Nsample_\text{leaf}}]$ and $\boldb_i \in \{0,1\}^{\Nsample}$. The shortest path distance between leaf $i$ and leaf $j$ can be represented as
\begin{align*}
    d_\calT(\boldx_i,\boldx_j)  &= \boldw^\top (\boldb_i + \boldb_j - 2 \boldb_i \circ \boldb_j).
\end{align*}

\begin{proof}
The shortest distance between leaf $i$ and leaf $j$ can be written as
\begin{align*}
    d_\calT(\boldx_i,\boldx_j)  &= \sum_{k \in \textnormal{Path}(\boldx_i,\boldx_j)} w_k \\
    &=\sum_{k \in \calS_i}w_k + \sum_{k' \in \calS_j} w_{k'} - 2 \sum_{k \in \calS_i \cap \calS_j} w_k\\
    &= \boldw^\top \boldb_i+ \boldw^\top\boldb_j - 2 \boldw^\top (\boldb_i \circ \boldb_j)\\
    &= \boldw^\top (\boldb_i + \boldb_j - 2 \boldb_i \circ \boldb_j),
\end{align*}
where $\calS_i$ includes the set of ancestor node indices of leaf $i$ and node index of leaf $i$. \proofend
\end{proof}
\end{prop}

\begin{figure*}[t]
\begin{minipage}{0.99\textwidth}
\begin{algorithm}[H]
    \centering
    \caption{Sliced Weight Estimation with Trees} \label{alg:alg-s}
   \begin{algorithmic}[1]
\STATE \textbf{Input:} The matrix $\boldX$, the regularization parameter $\lambda_1 \geq 0$, and a set of indices $\Omega$.
\FOR{$t = 1, 2, \ldots, T$}
\STATE \text{random.seed}($i$)
\STATE $\boldB_t := \text{QuadTree}(\boldX)$ or $\boldB_t := \text{ClusterTree}(\boldX)$
\STATE Compute $\boldz_{i,j}^{(t)}$ from $\boldB_t$ and $d(\boldx_i, \boldx_j), (i,j) \in \Omega$
\STATE  $\widehat{\boldw}_t :=\argmin_{\boldw \in \mathbbR_+^{\Nsample_t}} \sum_{(i,j) \in \Omega}(d(\boldx_i,\boldx_j) - \boldw^\top \boldz_{i,j}^{(t)})^2 + \lambda \|\boldw\|_1.$
\ENDFOR
\RETURN $\{(\boldB_t, \widehat{\boldw}_t)\}_{t =1}^T$
    \end{algorithmic}
    \vspace{0.07in}
\end{algorithm}
\end{minipage}
\vspace{-.1in}
\end{figure*}

Given the closed-form expression in Proposition \ref{prop:shortest_distance}, we formulate the edge weight estimation problem as follows. First, we assumed that the tree was obtained using a tree construction algorithm such as QuadTree \citep{indyk2003fast} and ClusterTree \citep{le2019tree}, and  we fixed the tree structure (i.e., fixed $\boldB$). We then propose the following optimization for learning $\boldw$:
\begin{align*}
    \widehat{\boldw} :=\argmin_{\boldw \in \mathbbR_+^\Nsample} \sum_{i,j = 1}^{m}  (d(\boldx_i,\boldx_j) - \boldw^\top \boldz_{i,j})^2 + \lambda \|\boldw\|_1,
\end{align*}
where $\boldz_{i,j} = \boldb_i + \boldb_j - 2 \boldb_i \circ \boldb_j \in \mathbbR_+^{N}$, $\|\boldw\|_1$ is the L1-norm, and $\lambda \geq 0$ is a regularization parameter. This optimization problem is convex with respect to $\boldw$ and we can easily solve the problem using an off-the-shelf nonnegative Lasso solver.  Note that $\boldz_{i,j}$ is a sparse vector with only $2H \ll n$ elements, where $H$ is the tree depth. Thus, we can store the entire $\boldz$ in a sparse format and efficiently solve the Lasso problem.  Moreover, thanks to the L1-regularization, we can make most of the weights zero; this corresponds to merging the nodes of the tree. Thus, with this optimization, we are not only able to estimate the tree's weight, but also compress its size. 

We further propose a subsampled variant of the regression method:
\begin{align}
     \label{eq:proposed-efficient}
     \widehat{\boldw} :=\argmin_{\boldw \in \mathbbR_+^\Nsample} \sum_{(i,j) \in \Omega}(d(\boldx_i,\boldx_j) - \boldw^\top \boldz_{i,j} )^2 + \lambda \|\boldw\|_1,
\end{align}
where $\Omega$ denotes a set of indices. In this paper, we randomly subsampled the indices. 

Although our proposed method is specifically designed for the 1-Wasserstein distance approximation, it can also be used to approximate any distance using shortest path distance on a tree.

\subsection{Tree-Slice extension}
Thus far, we have focused on estimating the distance using a single tree. However, the approximation performance is highly dependent on the tree that is fixed before the weight estimation. To address this issue,  we can use the tree-sliced Wasserstein distance \citep{le2019tree}:
\begin{align*}
    \bar{W}_{\calT}(\mu,\nu) = \frac{1}{T}\sum_{t =1}^T W_{\calT_t}(\mu,\nu),
\end{align*}
where $W_{\calT_t}(\mu,\nu)$ is TWD with the $t$th generated tree, and $T$ is the total number of trees. In previous studies, TWD outperformed approaches based on a single tree.  The matrix version of the tree-sliced Wasserstein distance can be written as
\begin{align*}
    \bar{W}_{\calT}(\mu,\nu) = \frac{1}{T}\sum_{t=1}^T \left\| \text{diag}(\boldw_t)\boldB_t(\bolda - \boldb) \right\|_1,
\end{align*}
where $\boldw_t$ and $\boldB_t$ are the weight and parameters of the $t$-th tree, respectively.

Here, we extend the single-tree weight estimation to multiple trees. To this end, we separately estimate $\widehat{\boldw}_t$ from each tree and then average the estimates:
\begin{align*}
    \bar{W}_{\calT}(\mu,\nu) = \frac{1}{T}\sum_{t=1}^T \left\| \text{diag}(\widehat{\boldw}_t)\boldB_t(\bolda - \boldb) \right\|_1,
\end{align*}
where 
\begin{align*}
     \widehat{\boldw}_t :=\argmin_{\boldw \in \mathbbR_+^{\Nsample_t}} \sum_{(i,j) \in \Omega}(d(\boldx_i,\boldx_j) - \boldw^\top \boldz_{i,j}^{(t)} )^2 + \lambda \|\boldw\|_1,
\end{align*}
$\boldz_{i,j}^{(t)} \in \mathbbR^{N_t}_+$ is computed from $\boldB_t$, where $N_t$ is the number of nodes in the $t$-th tree. We call the proposed methods with QuadTree and ClusterTree as Sliced-qTWD and Sliced-cTWD, respectively. Algorithm \ref{alg:alg-s} is the pseudocode for the weight estimation procedure using trees. If we set $T=1$, this corresponds to a single-tree case.

\subsection{Computational cost} The computational cost of the proposed algorithm consists of three parts: computing a tree, estimating the weight, and computing TWD. In many Wasserstein distance applications, such as document retrieval, it is necessary to compute the tree and weight only once. In this paper, we employed QuadTree and clusterTree, which are efficient methods of constructing trees.. For weight estimation, the computation of the Lasso problem is small. Overall, when the number of vectors is $10,000$ to $30,000$, the tree construction and weight estimation can be performed efficiently with a single CPU. For inference, TWD can be efficiently computed by using our tree-based approach, since its computational complexity is $O(\Nsample)$. For a computational comparison of TWD and other cases, please refer to \citep{takezawa2021supervised}. For the tree-sliced variant case, the computational cost of tree-sliced Wasserstein is $T$ times larger than that of a single tree case. 

\begin{table*}[t]
\small
\centering
\caption{The results of 1-Wasserstein distance approximation. We report the averaged mean absolute error (MAE), the averaged Pearson's correlation coefficients (PCC), and the averaged number of nodes used for computation, respectively. For tree-sliced variants, we empirically fix the number of sliced as $T=3$. For MAE and PCC computation, we computed the MAE and PCC between the 1-Wasserstein distance computed by EMD with Euclidean distance and the tree counterparts, and we run the algorithm 10 times by changing the random seed.  \label{tb:bccsport}}
\begin{tabular}{|c|c|c|c|c|c|c|c|c|c|}
\hline
 & \multicolumn{3}{c|}{Twitter}  & \multicolumn{3}{c|}{BBCSport} &\multicolumn{3}{c|}{Amazon} \\ 
Methods & MAE & PCC & Nodes & MAE & PCC & Nodes & MAE& PCC& Nodes\\ \hline
QuadTree                 & 0.486 & 0.702 & 5097.2 & 0.512 & 0.851 & 11463.1  & 0.573 & 0.596 & ~~34203.0\\ \hline
qTWD ($\lambda=10^{-3}$) & 0.053 & 0.880 & 4976.4 & 0.098 & 0.923 & 11016.4 & 0.072 & 0.789 & ~~32132.7\\ \hline
qTWD ($\lambda=10^{-2}$) & 0.053 & 0.879 & 4542.2 & 0.095 & 0.921 & ~~9968.5 & 0.067 & 0.783 & ~~28443.3\\ \hline
qTWD ($\lambda=10^{-1}$) & 0.051 & 0.873 & 4328.4 & 0.073 & 0.878 & ~~8393.8 & 0.051 & 0.732 & ~~16983.9\\ \hline \hline
ClusterTree              & 8.028 & 0.674 & 5717.5 & 5.842 & 0.799 & 12217.7  & 8.074 & 0.759 & ~~33430.0 \\ \hline
cTWD ($\lambda=10^{-3}$) & 0.039 & 0.885 & 5674.1 & 0.028 & 0.902 & 12145.3  & 0.036 & 0.785 & ~~32642.3 \\ \hline
cTWD ($\lambda=10^{-2}$) & 0.038 & 0.881 & 5481.0 & 0.023 & 0.894 & 11031.8  & 0.033 & 0.778 & ~~28701.8 \\ \hline
cTWD ($\lambda=10^{-1}$) & 0.036 & 0.867 & 4195.2 & 0.023 & 0.866 & ~~6680.7 & 0.028 & 0.768 & ~~~~9879.8\\ \hline\hline
Sliced-QuadTree                 & 0.463 & 0.796 & 15164.6 & 0.498  & 0.901 & 34205.7  & 0.552 & 0.759 & 102274.2 \\ \hline
Sliced-qTWD ($\lambda=10^{-3}$) & 0.053 & 0.890   & 14702.7  & 0.100 & 0.945 & 32447.4  & 0.072 & 0.838  & ~~94737.4 \\ \hline
Sliced-qTWD ($\lambda=10^{-2}$) & 0.052 & 0.892   & 13581.1  & 0.096  & 0.945  & 29966.5  & 0.067 & 0.849 & ~~85604.9\\ \hline
Sliced-qTWD ($\lambda=10^{-1}$) & 0.050 &  0.892   & 13026.3 & 0.073 & 0.925 & 25647.7 & 0.047 & 0.838 & ~~53689.9\\ \hline
Sliced-ClusterTree  & 7.989  & 0.804  & 17129.5  & 5.83 & 0.863  & 36644.2  & 8.071 & 0.822   & 100,251.3   \\ \hline
Sliced-cTWD ($\lambda=10^{-3}$) & 0.035  & 0.929  & 17002.6   & 0.026 & 0.948  & 36405.2 & 0.035 & 0.888 & ~~97891.5  \\ \hline
Sliced-cTWD ($\lambda=10^{-2}$) & 0.034  & 0.929  & 16414.7   & 0.021 & 0.943  & 33024.7 & 0.031  & 0.884 &  ~~86068.8\\ \hline
Sliced-cTWD ($\lambda=10^{-1}$) & 0.029  & 0.930  & 12565.3   & 0.018 & 0.924  & 19947.2 & 0.022 & 0.870  &  ~~29540.0\\ \hline
\end{tabular}
\vspace{-.1in}
\end{table*}

\begin{figure*}[t]
    \centering
    \begin{subfigure}[t]{0.3\textwidth}
        \centering
        \includegraphics[width=0.99\textwidth]{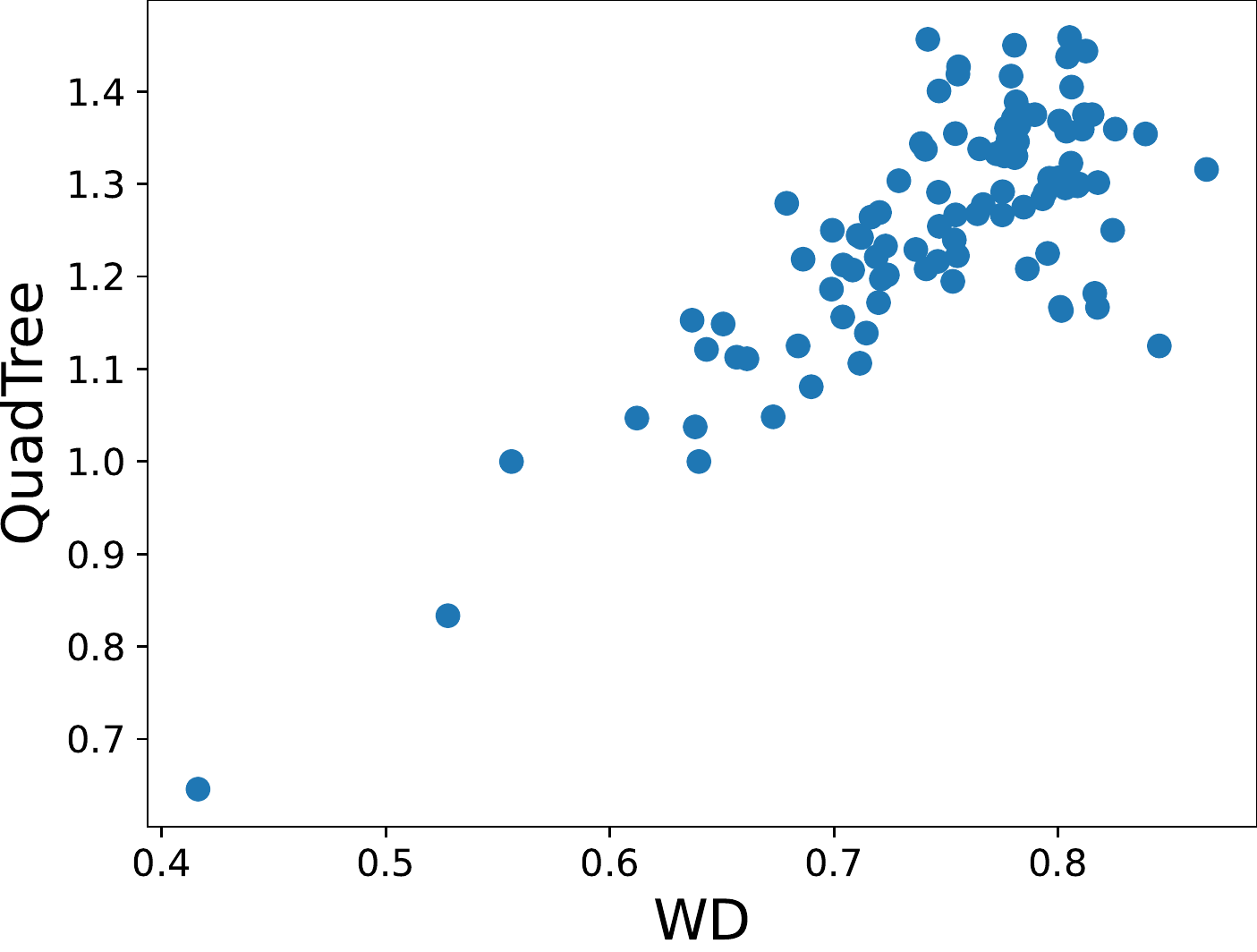}
        \caption{Twitter, QuadTree ($\rho=0.781$).}
    \end{subfigure}\quad
     \begin{subfigure}[t]{0.3\textwidth}
        \centering
        \includegraphics[width=0.99\textwidth]{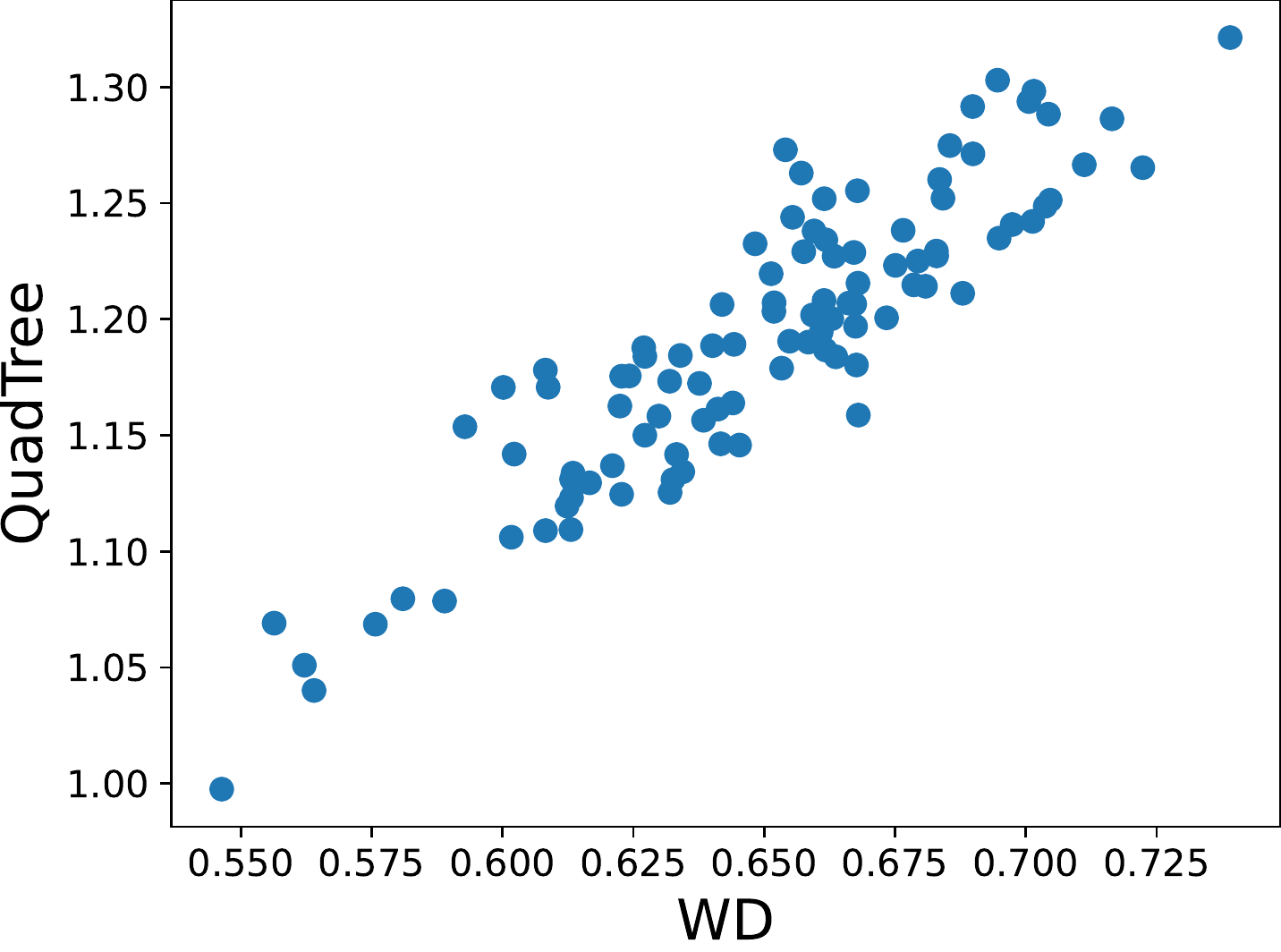}
        \caption{BBCSport, QuadTree ($\rho=0.904$).}
    \end{subfigure}\quad
    \begin{subfigure}[t]{0.3\textwidth}
        \centering
        \includegraphics[width=0.99\textwidth]{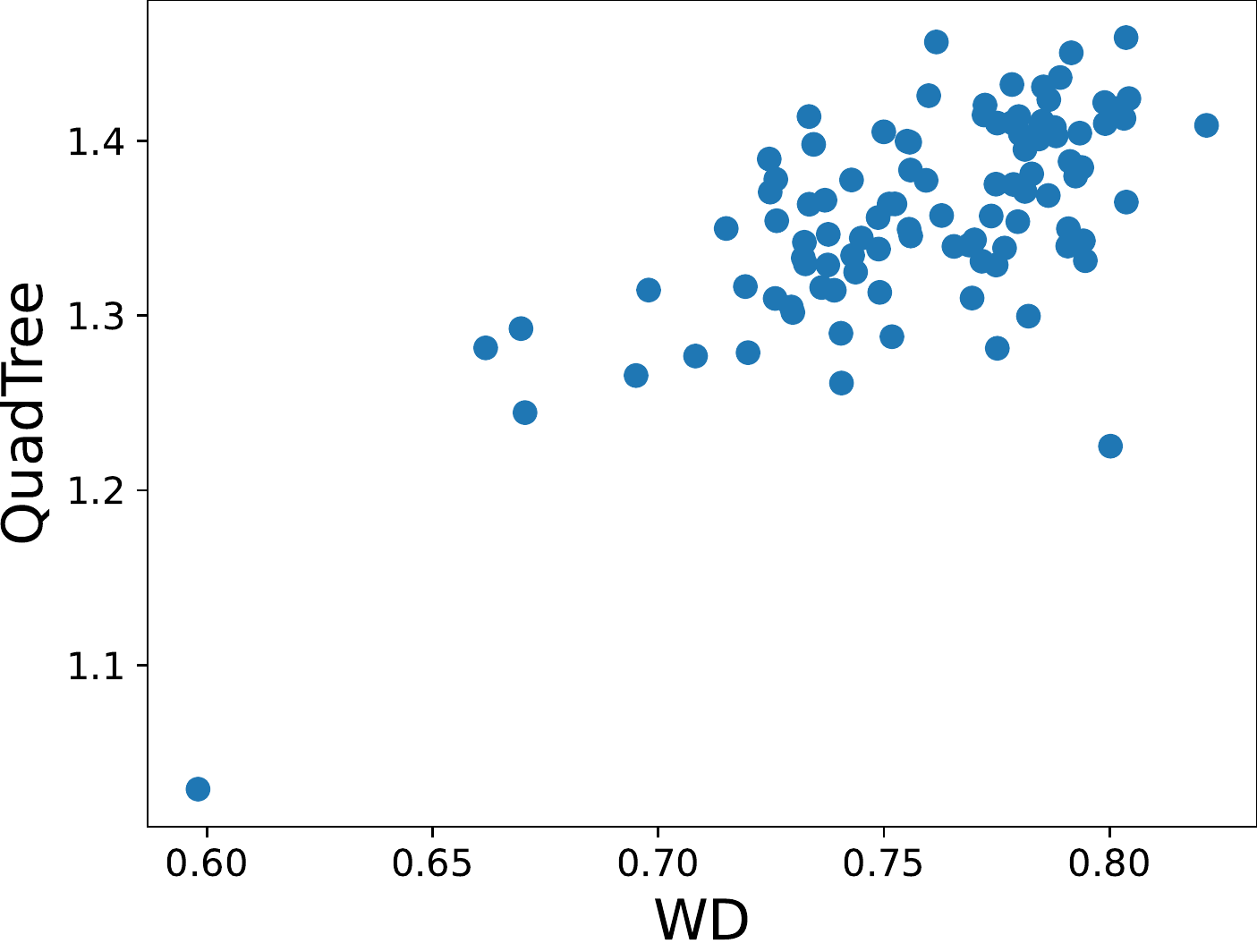}
        \caption{Amazon, QuadTree ($\rho=0.669$)}
    \end{subfigure}\quad
     \begin{subfigure}[t]{0.3\textwidth}
        \centering
\includegraphics[width=0.99\textwidth]{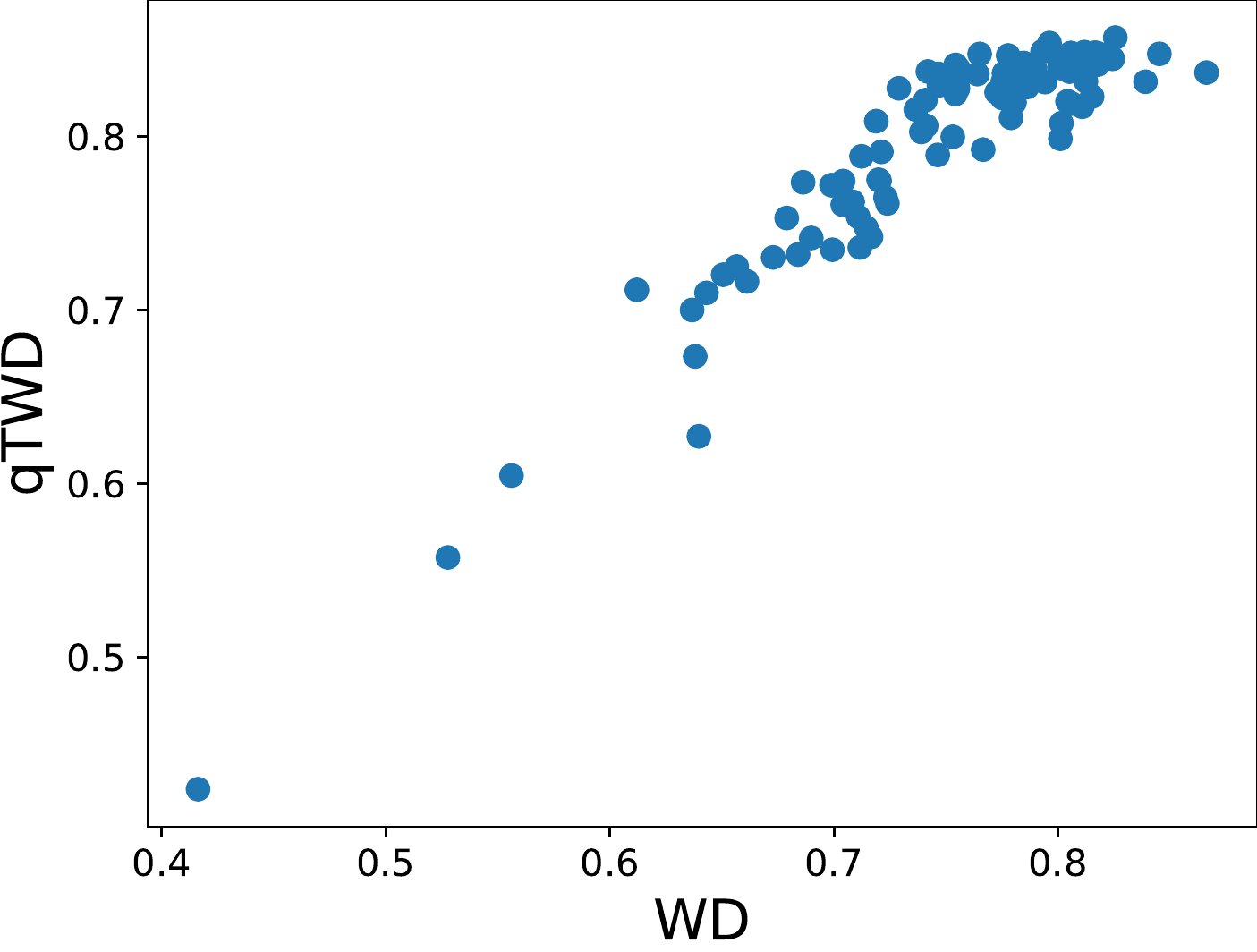}
        \caption{Twitter, qTWD ($\rho=0.930$).}
    \end{subfigure}\quad
        \begin{subfigure}[t]{0.3\textwidth}
        \centering
        \includegraphics[width=0.99\textwidth]{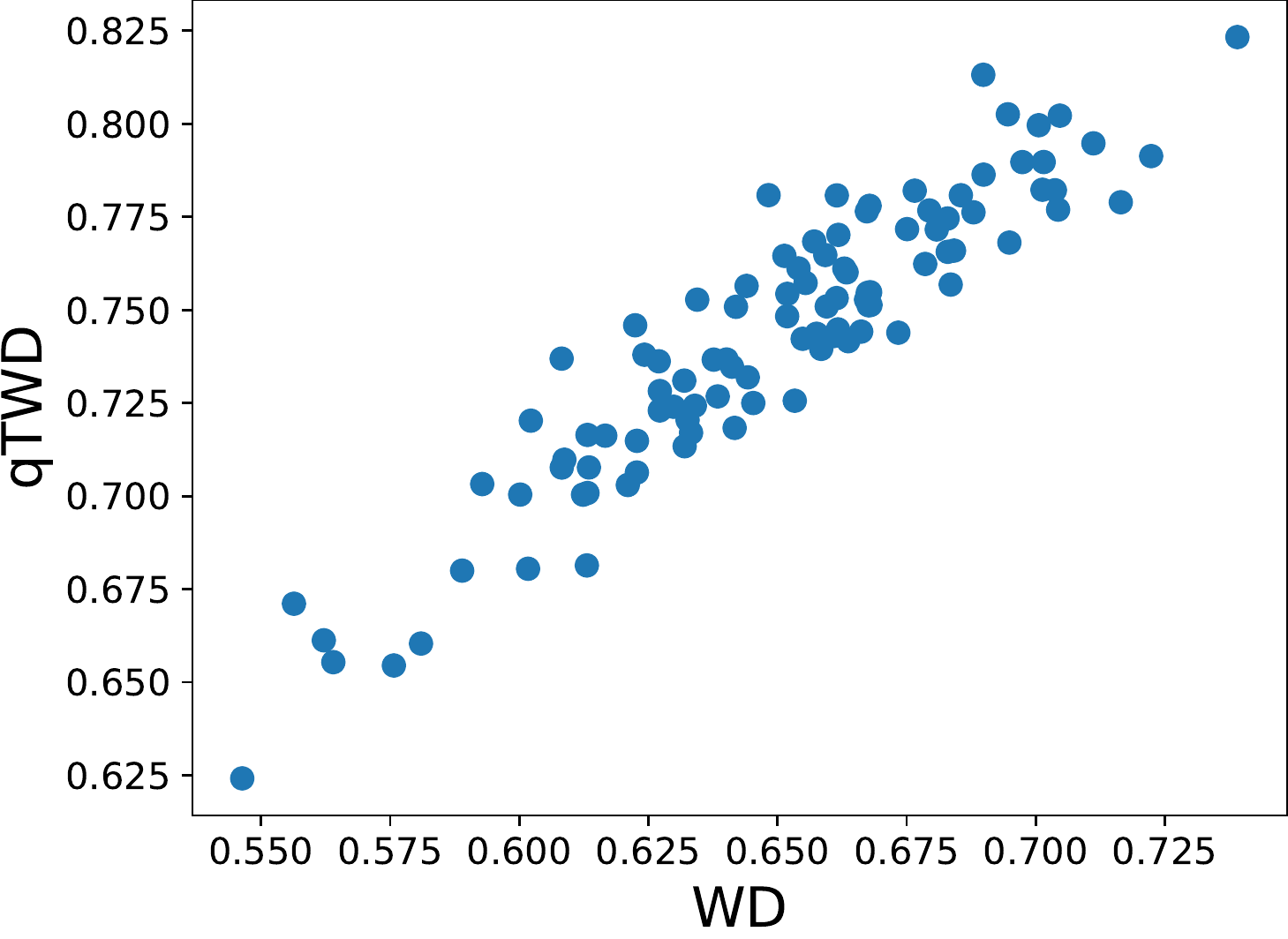}
        \caption{BBCSport, qTWD ($\rho=0.929$).}
    \end{subfigure}\quad
     \begin{subfigure}[t]{0.3\textwidth}
        \centering
\includegraphics[width=0.99\textwidth]{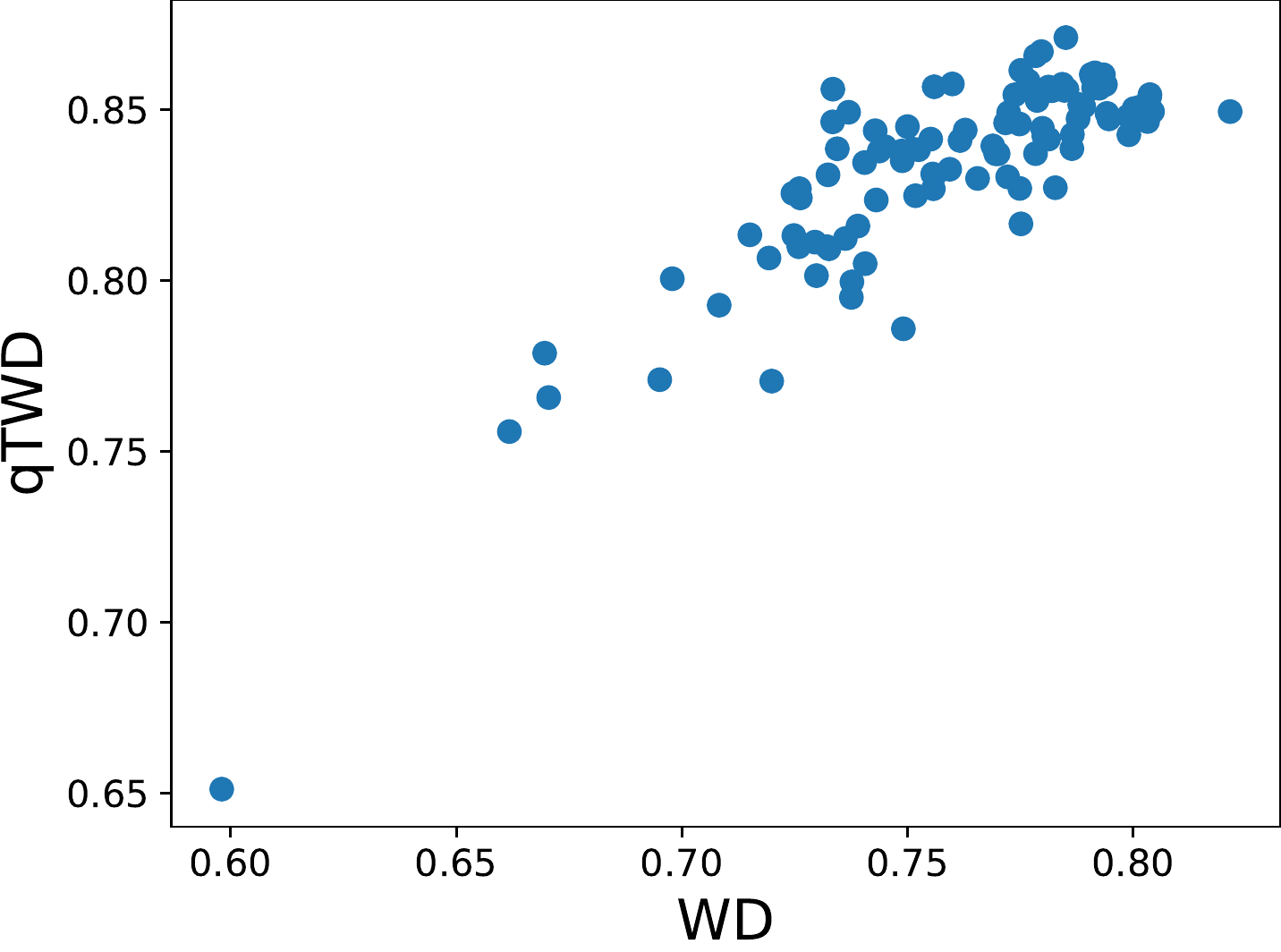}
        \caption{Amazon, qTWD ($\rho=0.848$).}
    \end{subfigure}\quad
            \begin{subfigure}[t]{0.3\textwidth}
        \centering
        \includegraphics[width=0.99\textwidth]{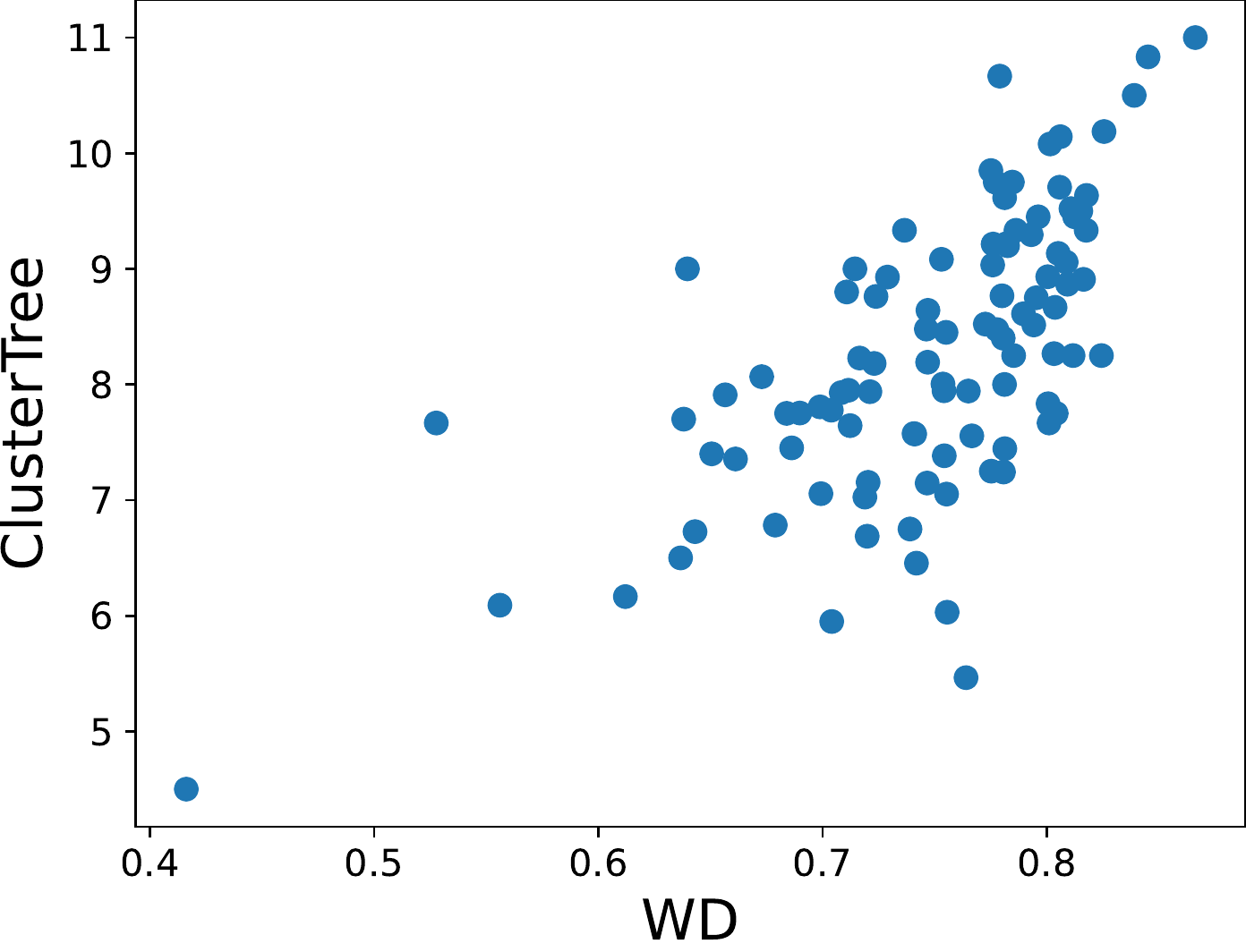}
        \caption{Twitter, ClusterTree ($\rho=0.651$).}
    \end{subfigure}\quad
     \begin{subfigure}[t]{0.3\textwidth}
        \centering
        \includegraphics[width=0.99\textwidth]{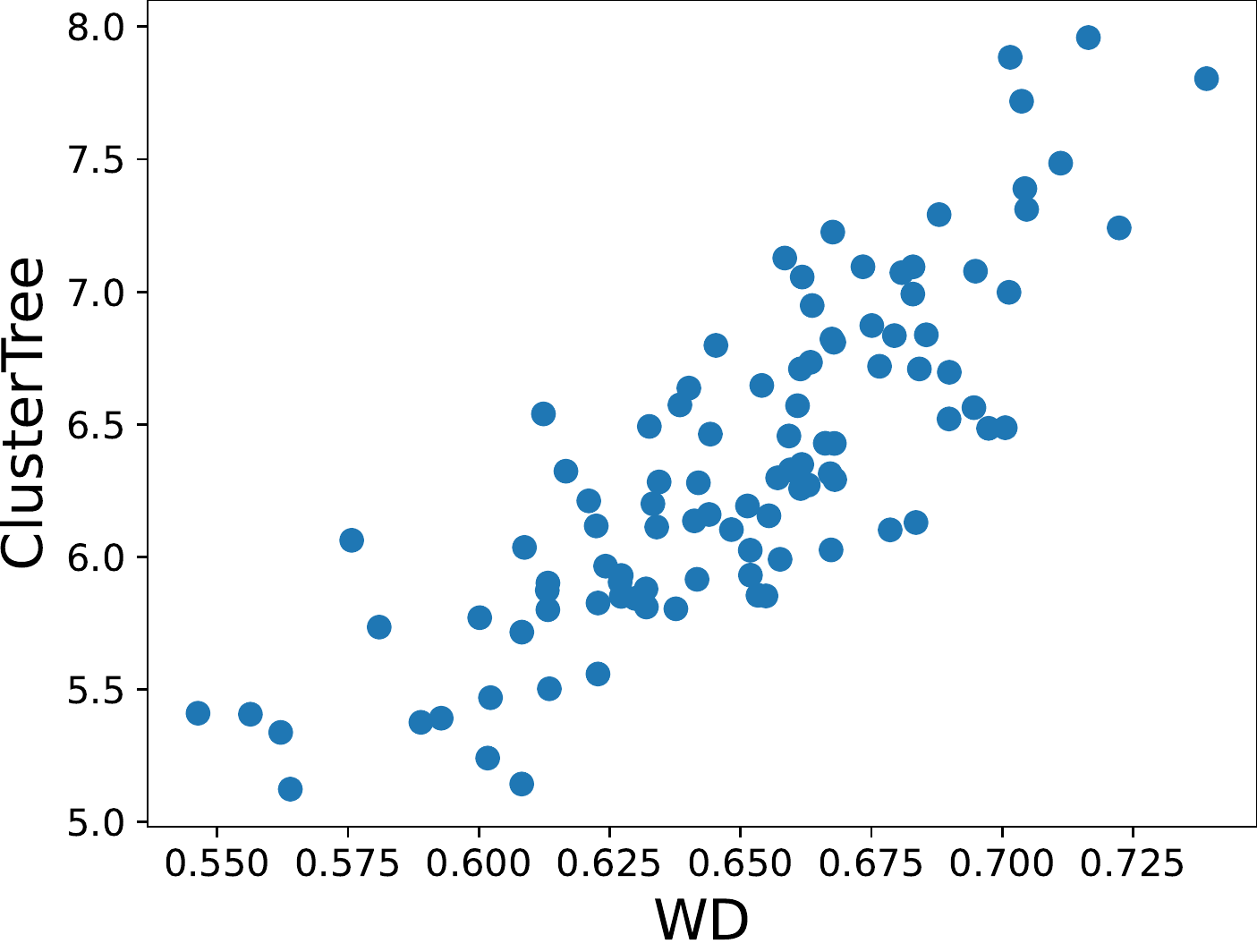}
        \caption{BBCSport, ClusterTree ($\rho=0.828$).}
    \end{subfigure}\quad
    \begin{subfigure}[t]{0.3\textwidth}
        \centering
        \includegraphics[width=0.99\textwidth]{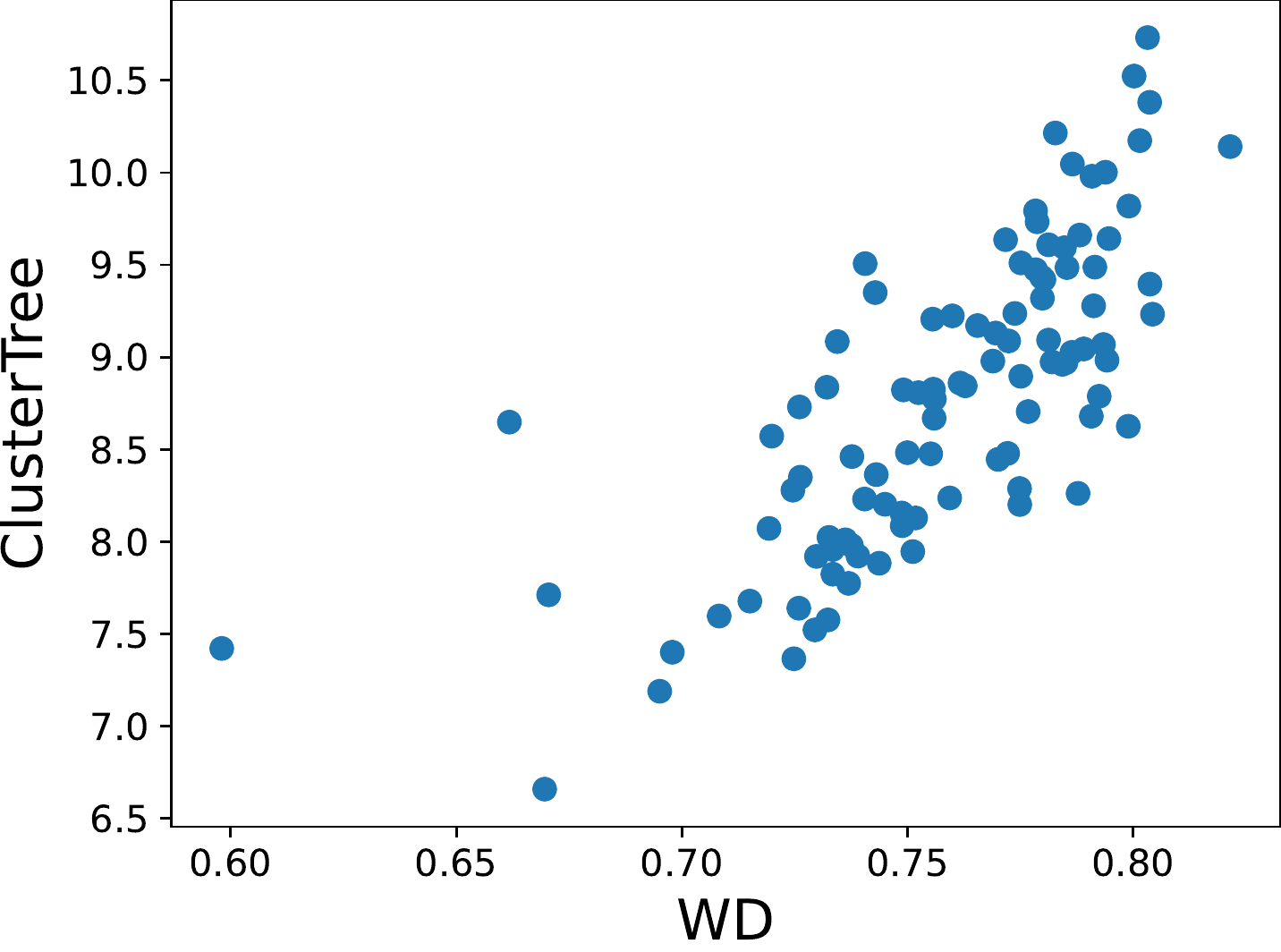}
        \caption{Amazon, ClusterTree ($\rho=0.747$)}
    \end{subfigure}\quad
     \begin{subfigure}[t]{0.3\textwidth}
        \centering
\includegraphics[width=0.99\textwidth]{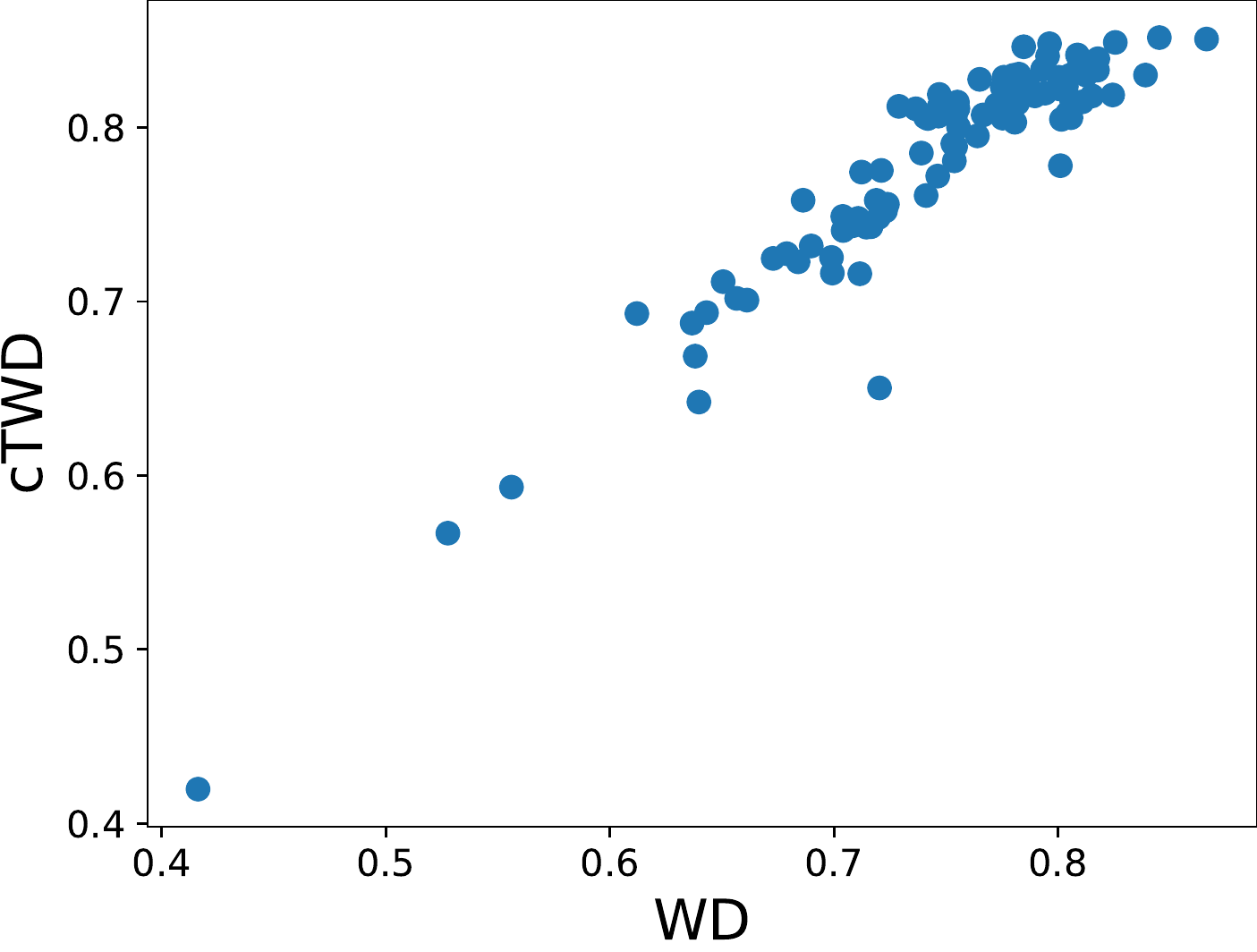}
        \caption{Twitter, cTWD ($\rho=0.942$).}
    \end{subfigure}\quad
        \begin{subfigure}[t]{0.3\textwidth}
        \centering
        \includegraphics[width=0.99\textwidth]{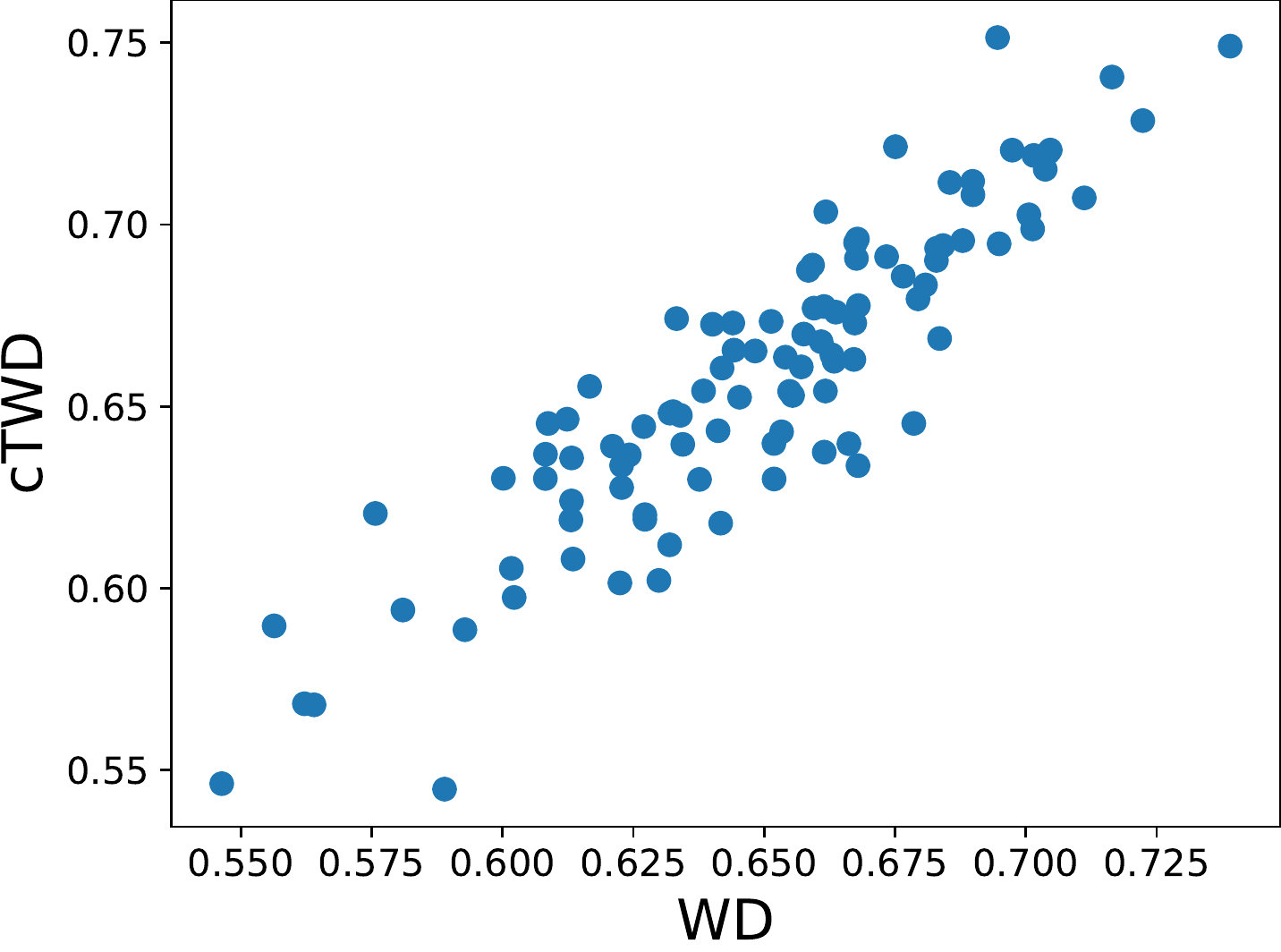}
        \caption{BBCSport, cTWD ($\rho=0.898$).}
    \end{subfigure}\quad
     \begin{subfigure}[t]{0.3\textwidth}
        \centering
\includegraphics[width=0.99\textwidth]{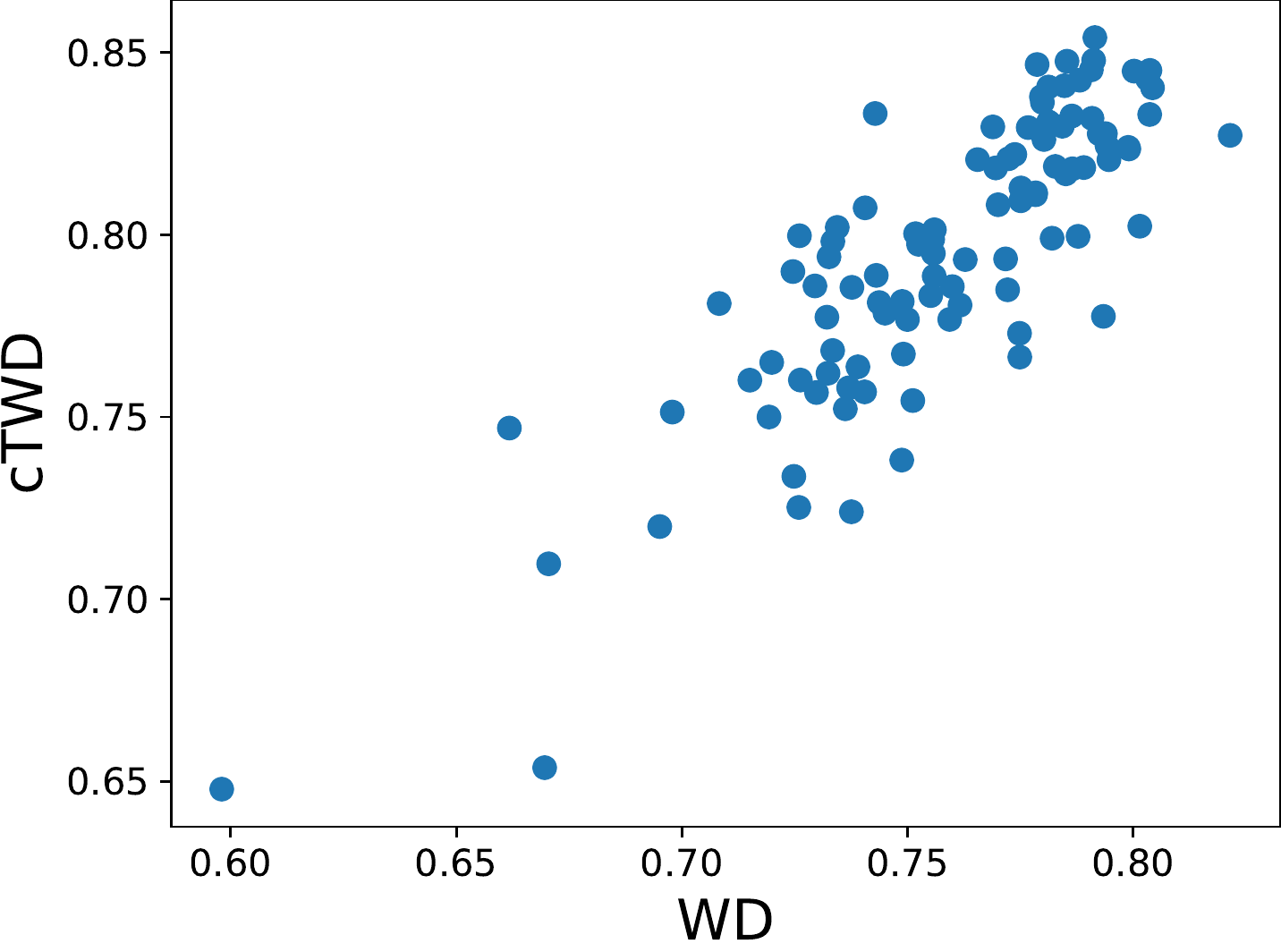}
        \caption{Amazon, cTWD ($\rho=0.841$).}
    \end{subfigure}\quad
    \caption{Scatter plots of the Twitter, BBCSport, and Amazon datasets.  \label{fig:scatter_quadtree}}
    \vspace{-.15in}
\end{figure*}

\section{Experiments}
In this section, we evaluate our proposed methods using Twitter, BBCSport, and Amazon datasets.

\subsection{Setup}
We evaluated the approximation ability of the proposed method. More specifically, we measured the error and correlation between the 1-Wasserstein distance with the Euclidean norm and tree-based methods. For the tree-based methods, we evaluated QuadTree \citep{indyk2003fast} and ClusterTree \citep{le2019tree}. For the 1-Wasserstein distance, we used the Python Optimal Transport (POT) package \footnote{\url{https://pythonot.github.io/index.html}}. For tree methods, we first construct a tree using an entire word embedding vector $\boldX$ and then compute the Wasserstein distance with the tree. For the proposed methods, we selected the regularization parameter from $\{10^{-3},10^{-2},10^{-1}\}$. qTWD and cTWD are both proposed methods, and qTWD and cTWD employ QuadTree and ClusterTree, respectively.  We compared all the methods using the Twitter, BBCSport, and Amazon datasets \footnote{\url{https://github.com/gaohuang/S-WMD}}. We further evaluated the group-sliced TWD using QuadTree and ClusterTree. We randomly subsample 100 document pairs and computed the mean absolute error (MAE) and Pearson's correlation coefficient (PCC) between the 1-Wasserstein distance computed by EMD with Euclidean distance and the tree counterparts. Moreover, we reported the number of nonzero tree weights. In this experiment, we set the number of slices to $T=3$ and the regularization parameters $\lambda = \{10^{-3},10^{-2},10^{-1}\}$.  We used SPAMS to solve Lasso problems \footnote{\url{http://thoth.inrialpes.fr/people/mairal/spams/}}. For all methods, we ran the algorithm ten times by changing the random seed.  We evaluated all methods using Xeon CPU E5-2690 v4 (2.60 GHz) and Xeon CPU E7-8890 v4 (2.20GHz). Note that since the number of vectors $\boldx_i$ tends to be large for real-world applications, we only evaluated the subsampled training method.

\begin{figure*}[t]
    \centering
        \begin{subfigure}[t]{0.32\textwidth}
        \centering
        \includegraphics[width=0.99\textwidth]{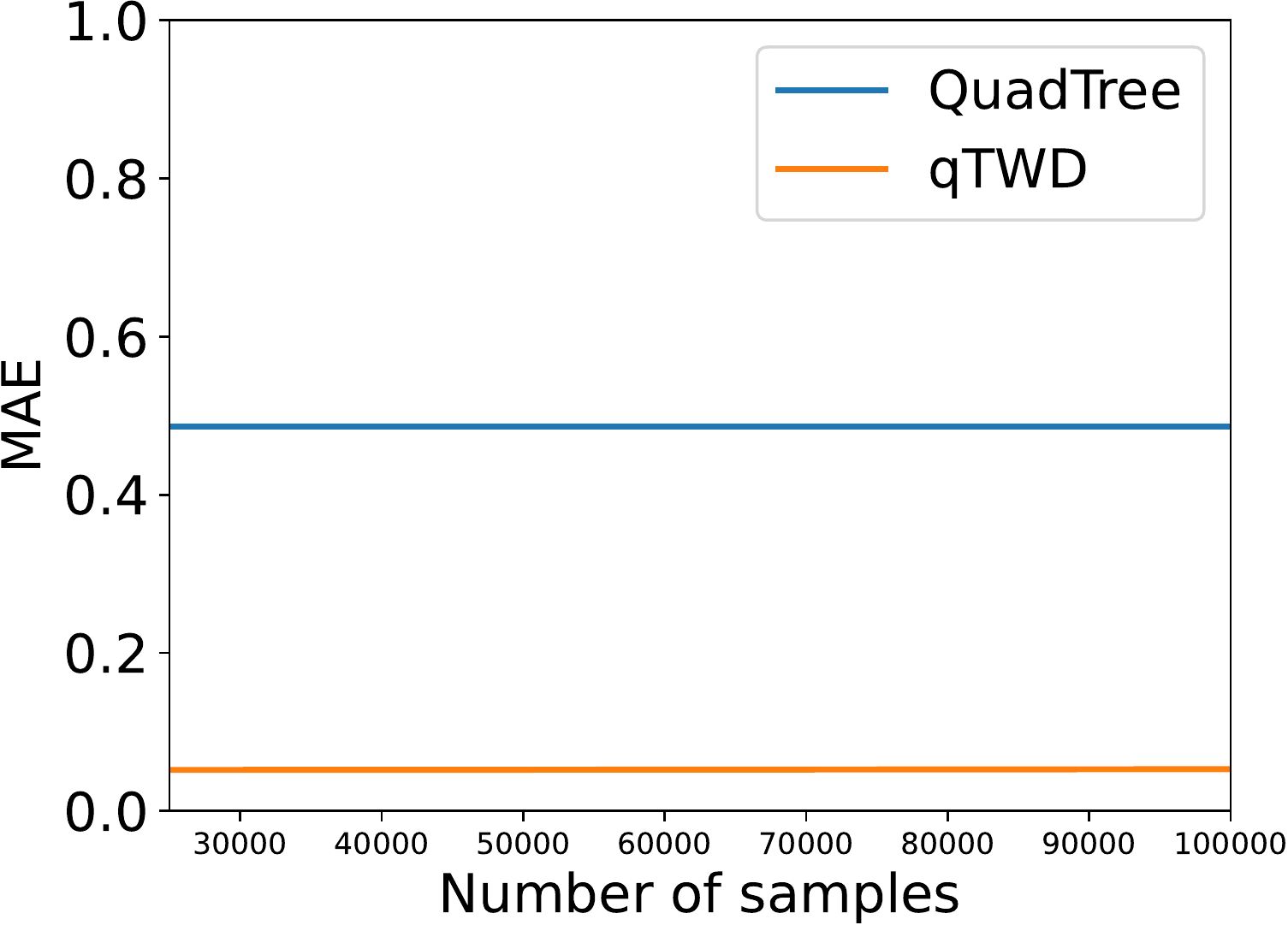}
        \caption{Twitter, qTWD.}
    \end{subfigure}\quad
     \begin{subfigure}[t]{0.32\textwidth}
        \centering
        \includegraphics[width=0.99\textwidth]{FIG/twitter_proposed_quad_sample_MAE.pdf}
        \caption{BBCSport, qTWD.}
    \end{subfigure}
        \begin{subfigure}[t]{0.32\textwidth}
        \centering
        \includegraphics[width=0.99\textwidth]{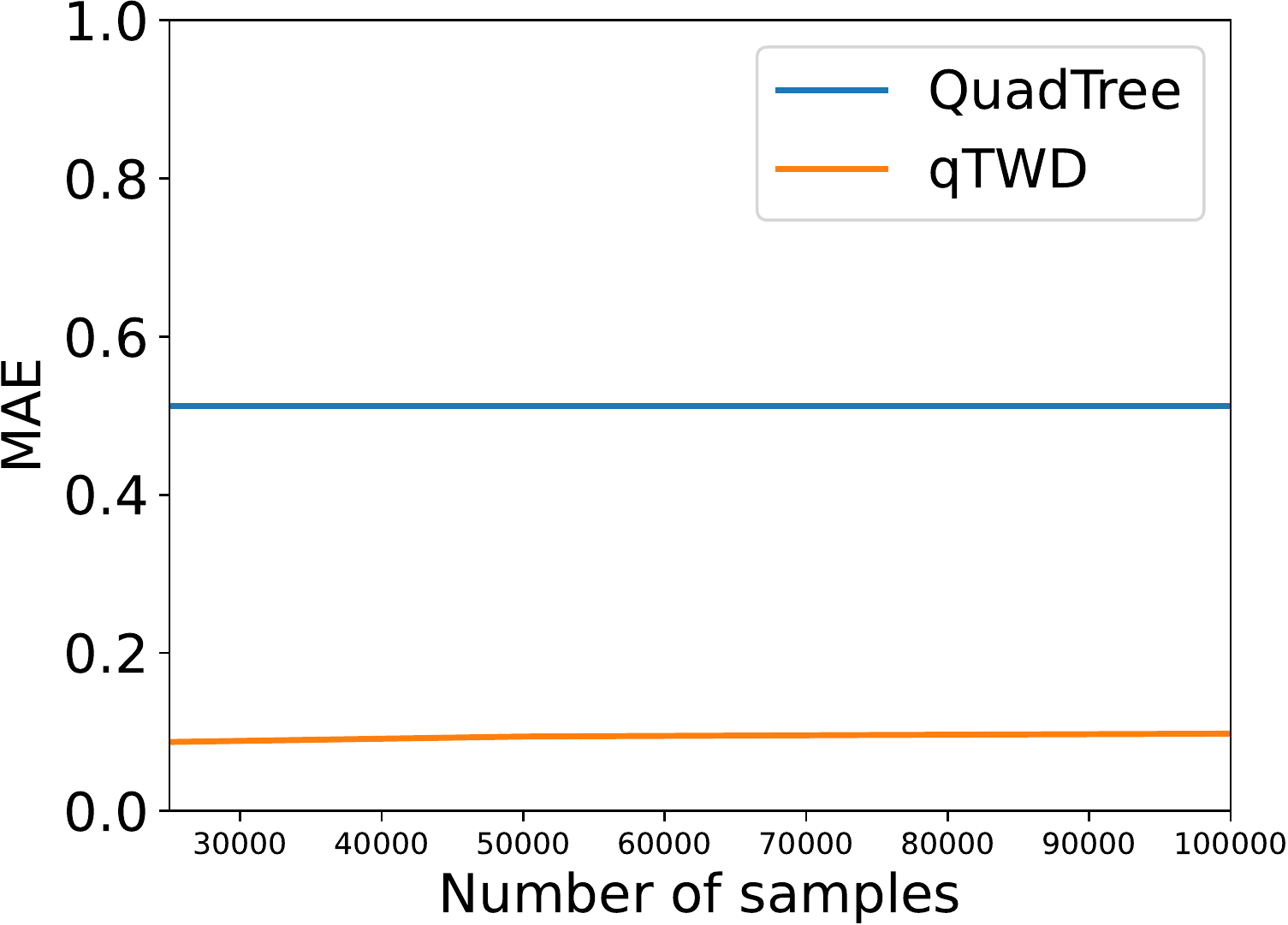}
        \caption{Amazon, qTWD.}
    \end{subfigure}\quad
        \begin{subfigure}[t]{0.32\textwidth}
        \centering
        \includegraphics[width=0.99\textwidth]{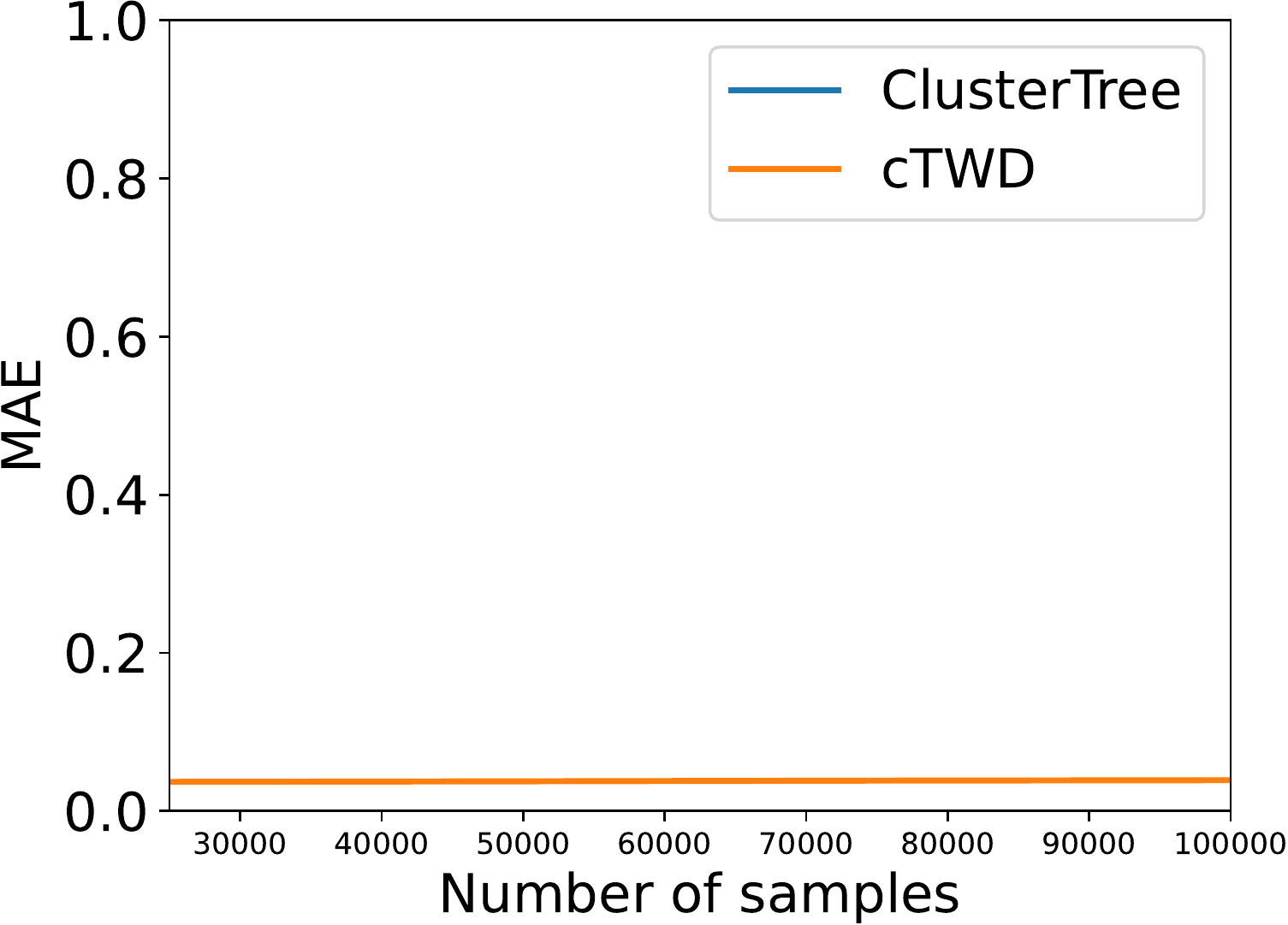}
        \caption{Twitter, cTWD.}
    \end{subfigure}\quad
     \begin{subfigure}[t]{0.32\textwidth}
        \centering
        \includegraphics[width=0.99\textwidth]{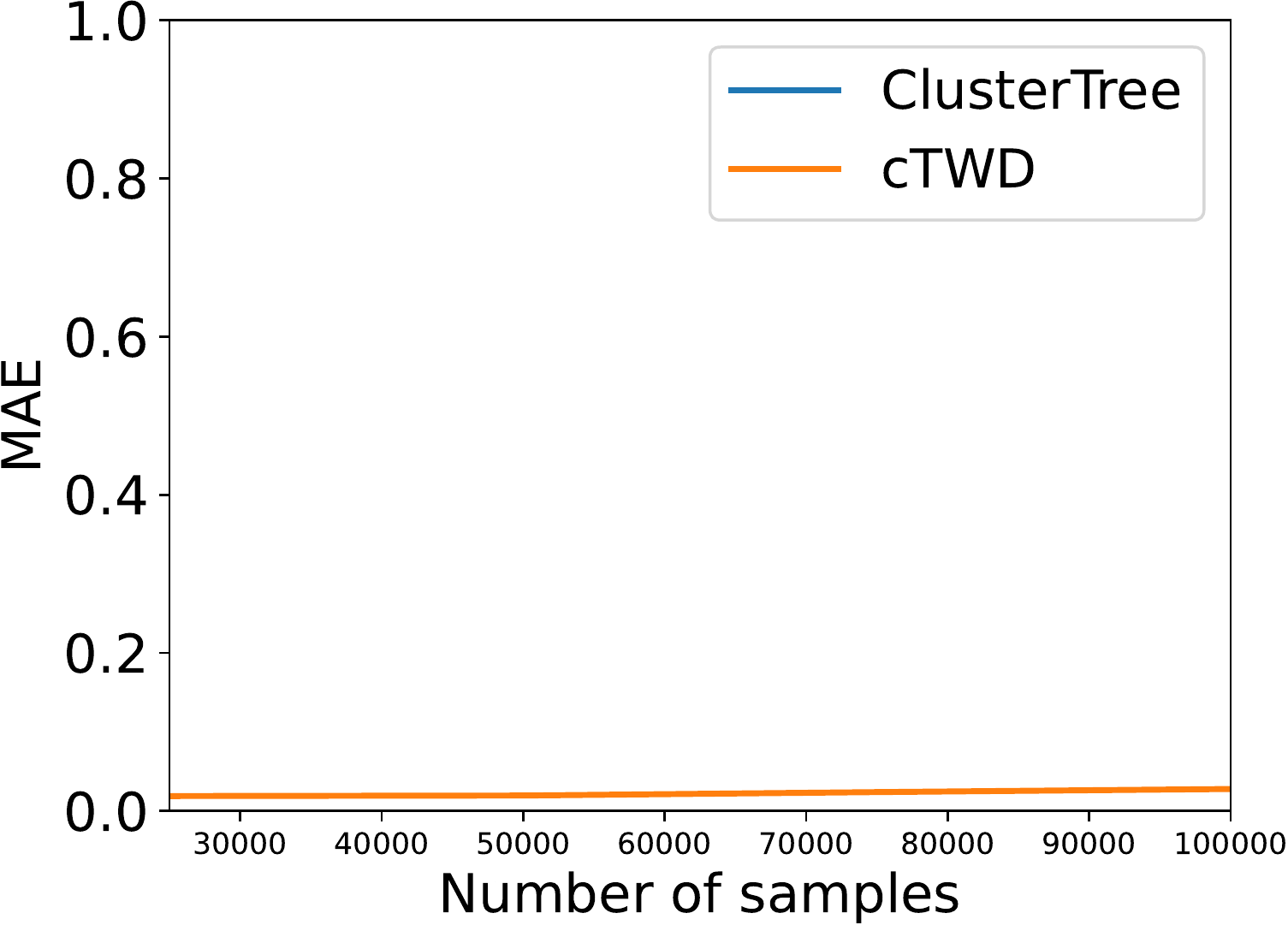}
        \caption{BBCSport, cTWD.}
    \end{subfigure}
        \begin{subfigure}[t]{0.32\textwidth}
        \centering
        \includegraphics[width=0.99\textwidth]{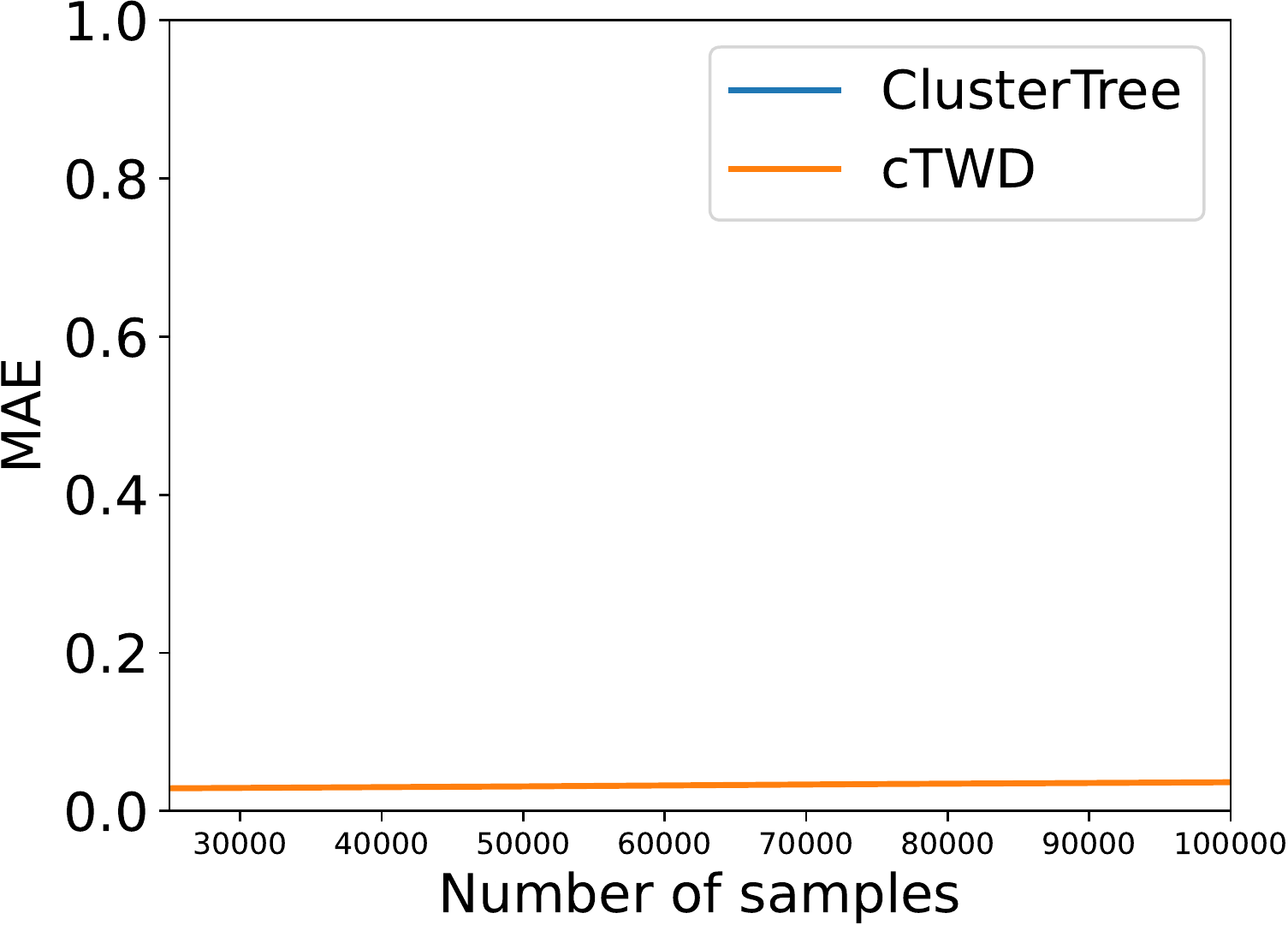}
        \caption{Amazon, cTWD.}
    \end{subfigure}\quad
         \begin{subfigure}[t]{0.32\textwidth}
        \centering
        \includegraphics[width=0.99\textwidth]{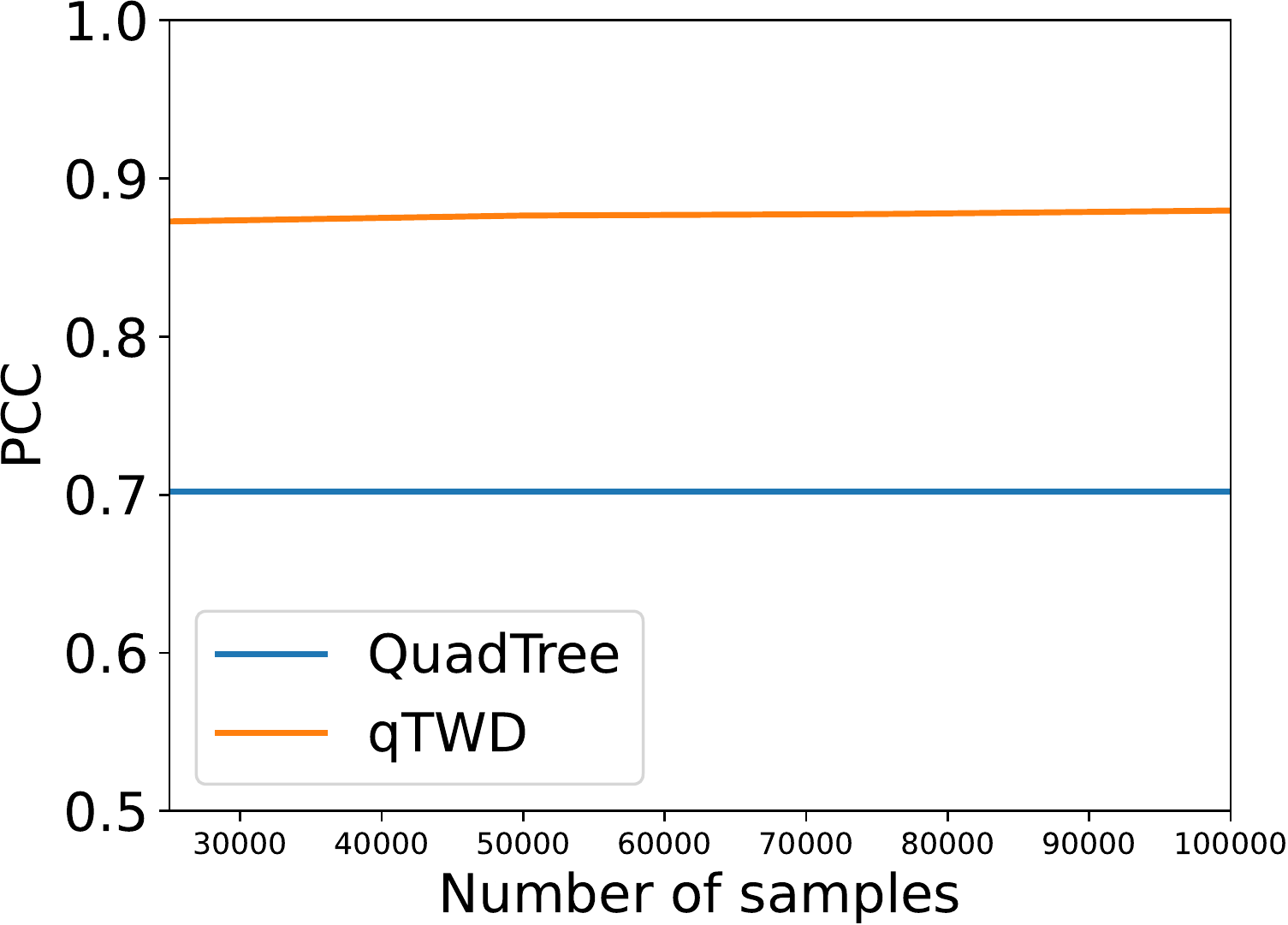}
        \caption{Twitter, qTWD.}
    \end{subfigure}
        \begin{subfigure}[t]{0.32\textwidth}
        \centering
        \includegraphics[width=0.99\textwidth]{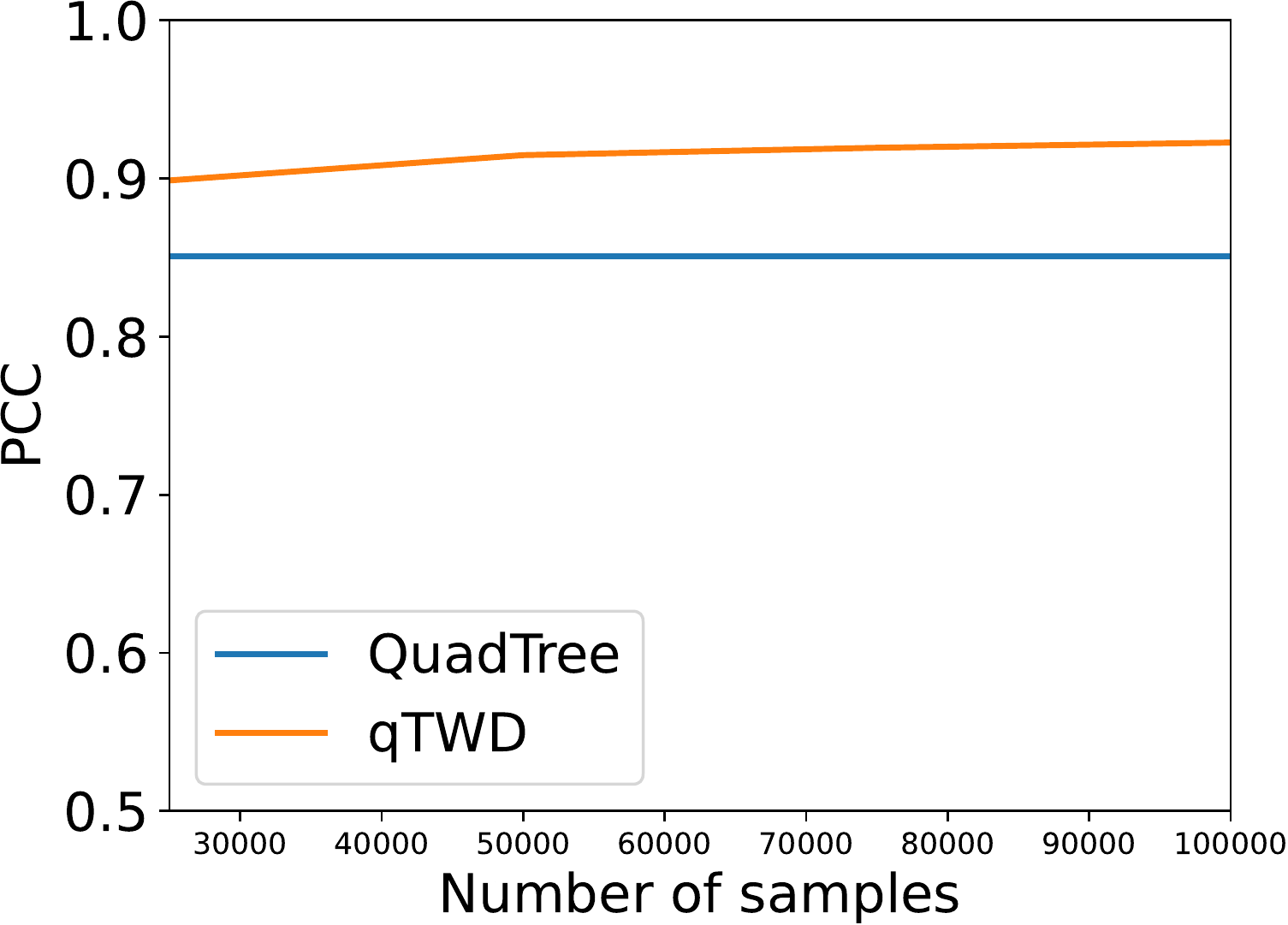}
        \caption{BBCSport, qTWD.}
    \end{subfigure}\quad
     \begin{subfigure}[t]{0.32\textwidth}
        \centering
        \includegraphics[width=0.99\textwidth]{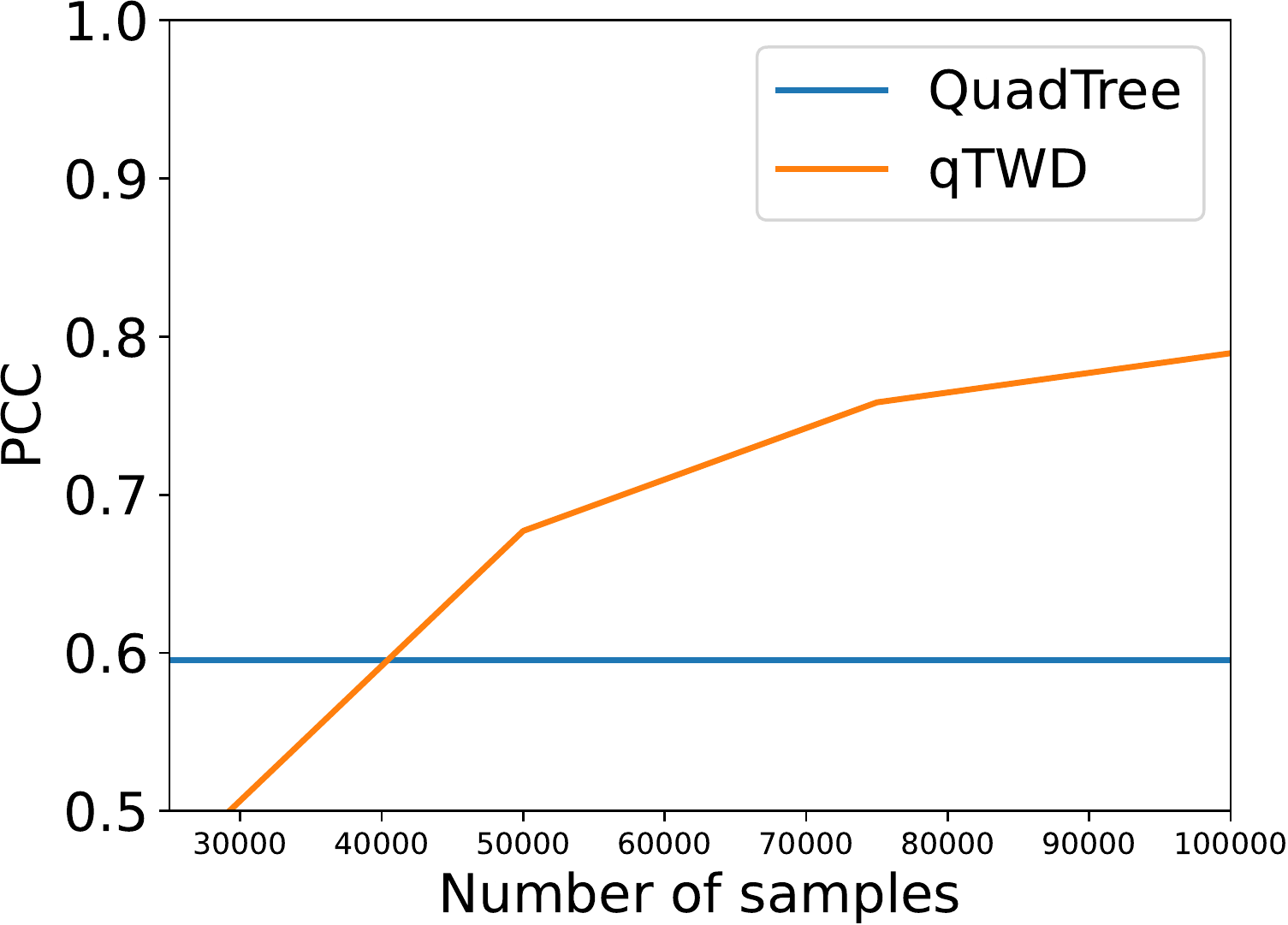}
        \caption{Amazon, qTWD.}
    \end{subfigure}
     \begin{subfigure}[t]{0.32\textwidth}
        \centering
        \includegraphics[width=0.99\textwidth]{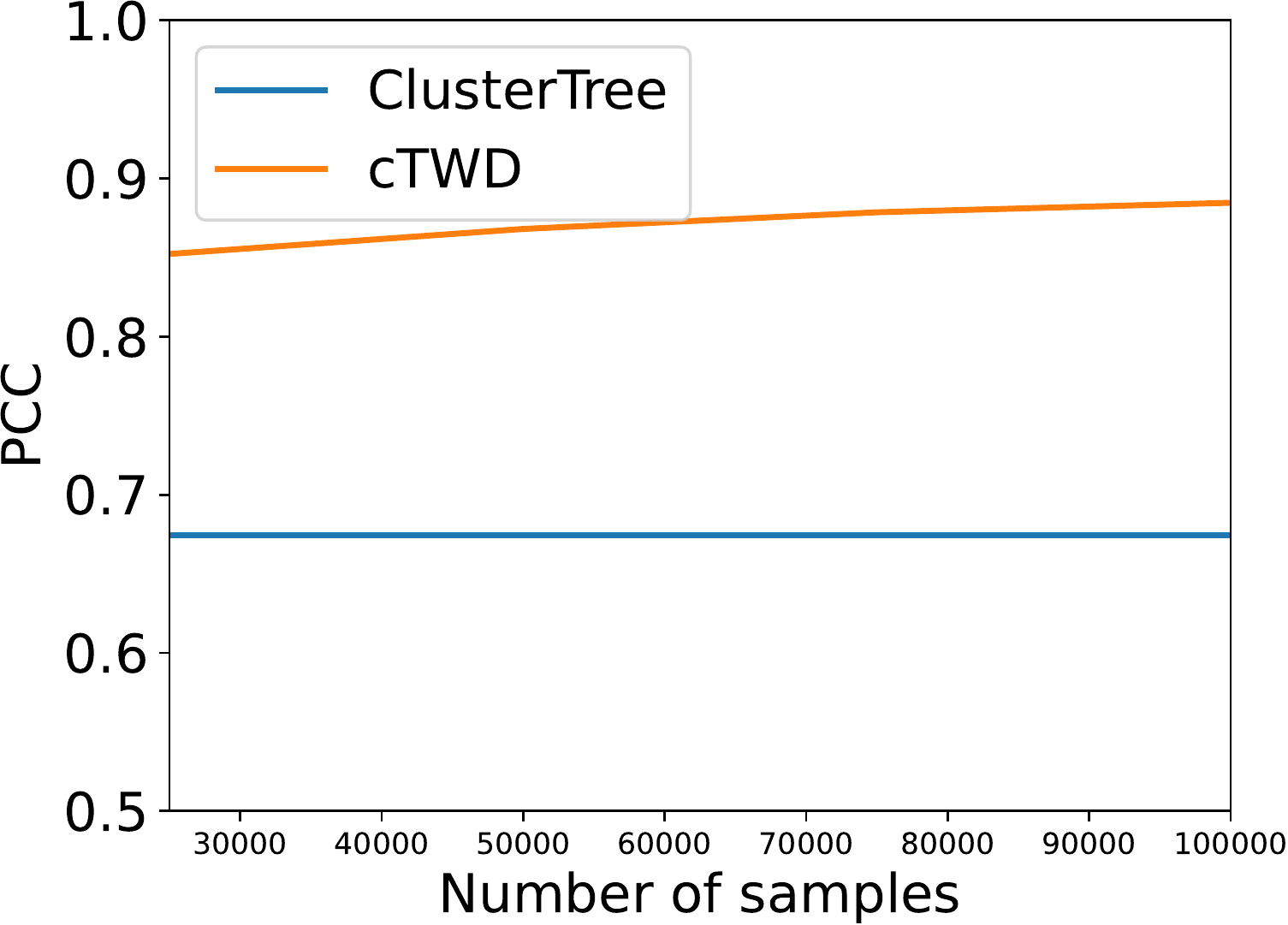}
        \caption{Twitter, cTWD.}
    \end{subfigure}
        \begin{subfigure}[t]{0.32\textwidth}
        \centering
        \includegraphics[width=0.99\textwidth]{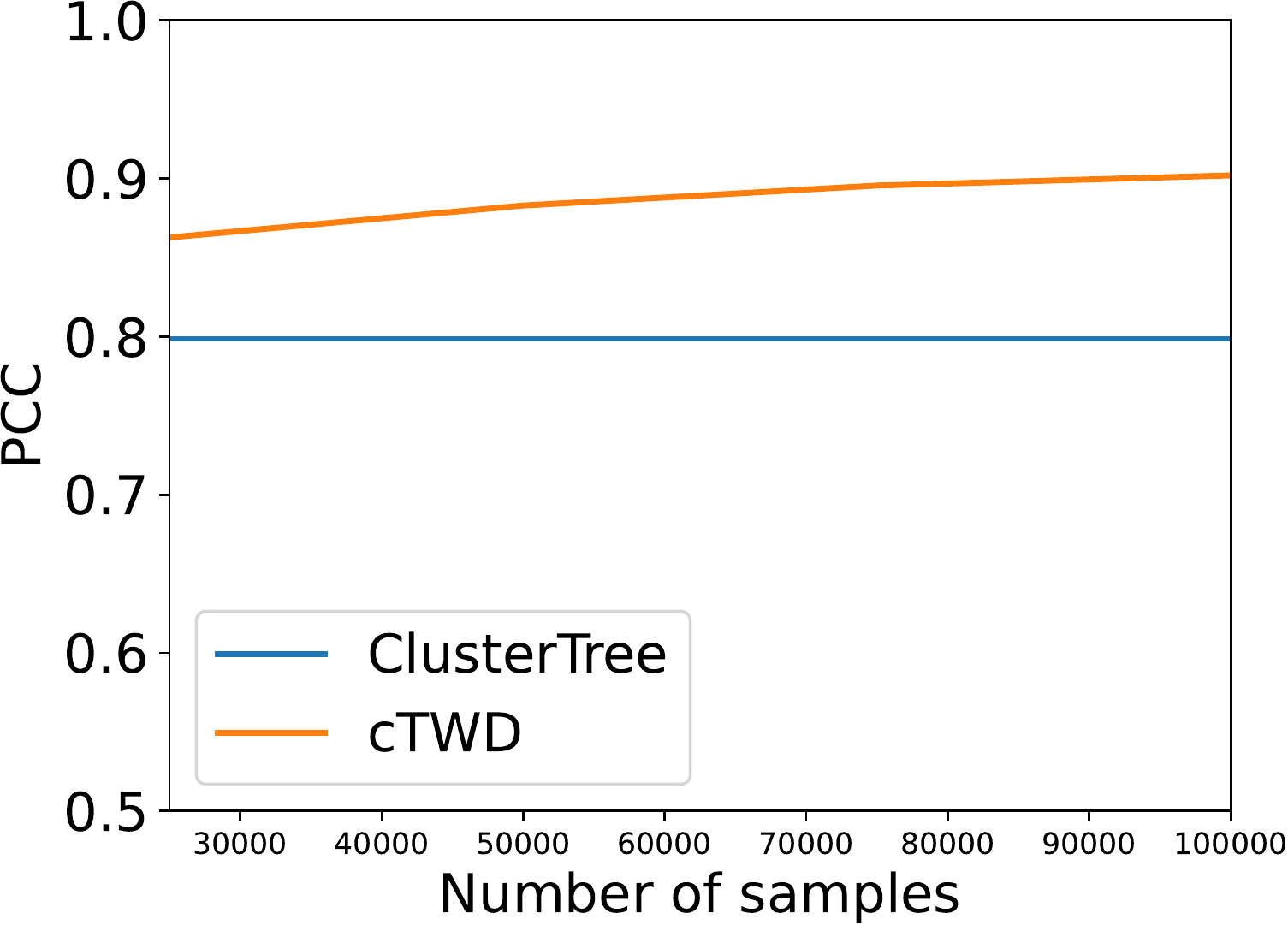}
        \caption{BBCSport, cTWD.}
    \end{subfigure}\quad
     \begin{subfigure}[t]{0.32\textwidth}
        \centering
        \includegraphics[width=0.99\textwidth]{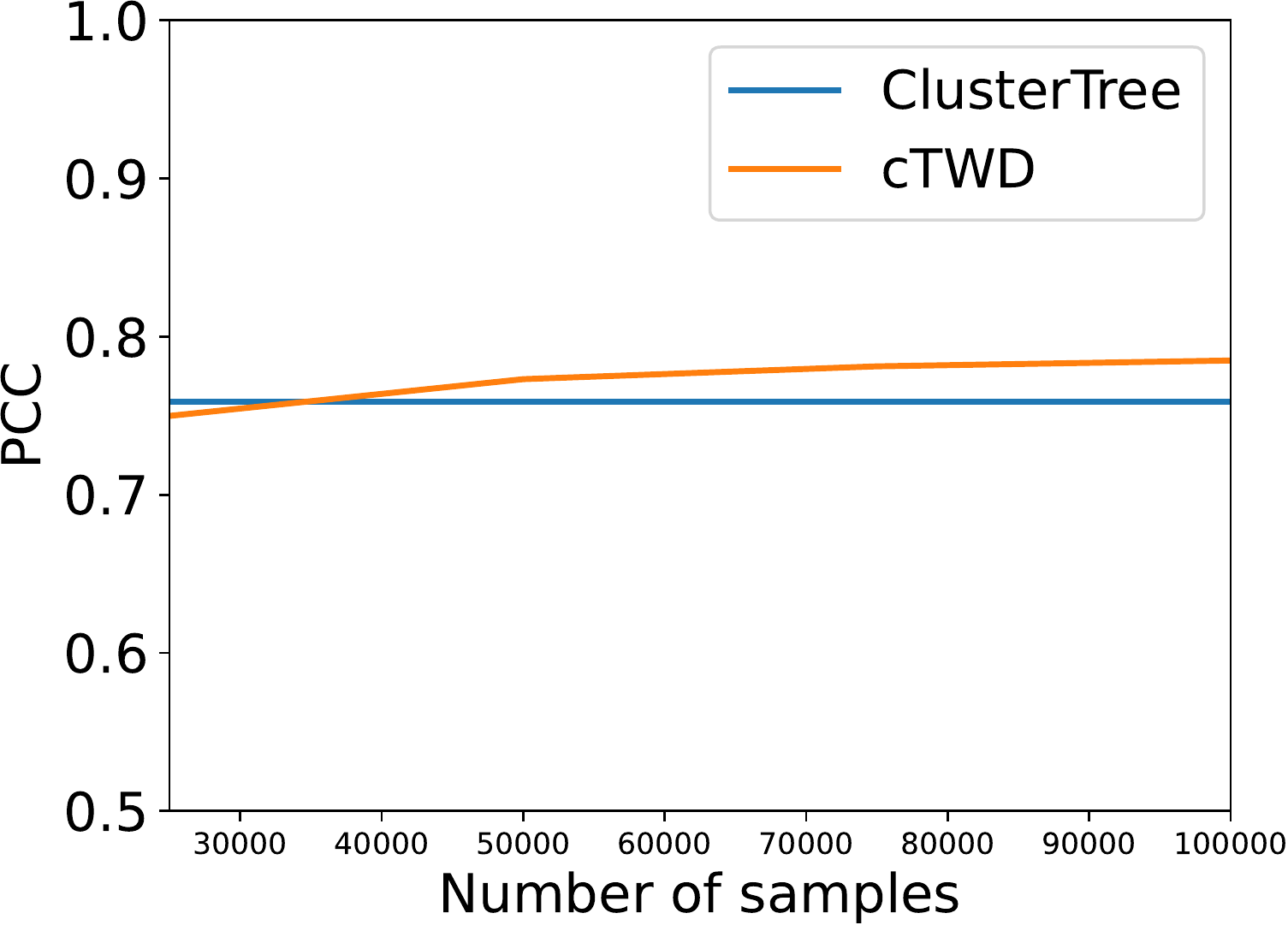}
        \caption{Amazon, cTWD.}
    \end{subfigure}
    \caption{MAE and PCC with respect to the number of training samples. In this experiments, we computed MAE and PCC for $m \in \{25,000,50,000, 75,000, 100,000\}$.  \label{fig:sample}}
\end{figure*}

\subsection{Results}
Table \ref{tb:bccsport} presents the experimental results for the Twitter, BBCSport, and Amazon datasets. As shown, qTWD and cTWD can obtain a small MAE and accurately approximate the original 1-Wasserstein distance. In contrast, QuadTree and ClusterTree had larger MAE values than the proposed methods. Although QuadTree has a theoretical guarantee, the MAE of QuadTree is one order of magnitude larger than that of the proposed method. For ClusterTree, since ClusterTree is based on clustering and the tree construction is independent of the scaling of vectors, it cannot guarantee a good approximation of 1-Wasserstein. However, our proposed method can approximate 1-Wasserstein even if we use ClusterTree. For PCC, qTWD and cTWD outperformed the vanilla QuadTree and ClusterTree if we set a small $\lambda$. Moreover, even when using half of the nodes (i.e., $\lambda = 10^{-1}$), PCCs of qTWD and cTWD were comparable to QuadTree and ClusterTree. Figure \ref{fig:scatter_quadtree} shows scatter plots for each method. We can see that all methods correlate with the 1-Wasserstein distance. However, it is evident that the proposed method has a similar scale to that of the original WD, whereas QuadTree and ClusterTree have larger values than the original WD. Moreover, Figure \ref{fig:sample} shows the effect of the training sample size used for training Eq. \eqref{eq:proposed-efficient}. In these experiments, the number of samples was changed to $25,000$, $50,000$, $75,000$, and $100,000$. We found that PCC could be improved by using more samples. In contrast, the MAE is small, even if we use a relatively small number of samples.

 With the tree-sliced, as reported in \citep{le2019tree}, Sliced-QuadTree and Sliced-ClusterTree outperform their non-sliced counterparts. Although there is no gain in MAE, our proposed methods, Sliced-qTWD and Sliced-cTWD can significantly improve the PCC values. Furthermore, because we use distance to retrieve similar documents, we can obtain similar results using the Wasserstein distance. More interestingly, for the Amazon dataset, Sliced-cTWD ($\lambda=10^{-1}$) has a PCC value of 0.870 with 29,540 nodes, whereas cTWD ($\lambda=10^{-3}$) has a PCC value of 0.785 with 32,642.3 nodes. Thus, Sliced-cTWD significantly outperformed cTWD. Thus, to obtain a low approximation error, one strategy is to use the sliced version and prune unimportant nodes using Lasso.

\section{Conclusion}
In this paper, we consider approximating the 1-Wasserstein distance with trees. More specifically, we first showed that the 1-Wasserstein distance approximation can be formulated as a distance-approximation problem. Then, we proposed a non-negative Lasso-based optimization technique for learning the weights of a tree. Because the proposed method is convex, a globally optimal solution can be obtained. Moreover, we proposed a weight estimation procedure for a tree-sliced variant of Wasserstein distance. Through experiments, we show that the proposed weight estimation method can significantly improve the approximation performance of the 1-Wasserstein distance for both QuadTree and clustered tree cases. Owing to L1-regularization, we can compress the tree size without losing approximation performance.  

\bibliography{main}
\bibliographystyle{plainnat}

\newpage
\appendix
\onecolumn

\section{Proof of Proposition \ref{prop:expressive}} \label{sec:proof-expressive}


Let $\boldPi^* = \boldPi^*(\mu, \nu)$. If there exists $(i, j, k)$ such that $\pi_{ij} > 0$ and $\pi_{jk} > 0$, \begin{align}
    \pi'_{pq} = \begin{cases}
    \pi_{ij} - \min(\pi_{ij}, \pi_{jk}) & (p, q) = (i, j) \\
    \pi_{jk} - \min(\pi_{ij}, \pi_{jk}) & (p, q) = (j, k) \\
    \pi_{ik} + \min(\pi_{ij}, \pi_{jk}) & (p, q) = (i, k) \\
    \pi_{pq} & \text{otherwise}\\
    \end{cases}
\end{align}
is no worse than $\boldPi^*$ because $d(i, j) + d(j, k) \ge d(i, k)$ due to the triangle inequality. We assume no tuples $(i, j, k)$ exists such that $\pi_{ij} > 0$ and $\pi_{jk} > 0$ without loss of generality. If there exist $i_1, i_2, \cdots, i_{2n}, i_{2n+1} = i_1$ such that $\boldP_{i_1 i_2} > 0, \boldP_{i_3 i_2} > 0, \boldP_{i_3 i_4} > 0, \cdots, \boldP_{i_1 i_{2n}} > 0$, either \begin{align}
    \pi'_{pq} = \begin{cases}
    \pi_{i_{2k+1} i_{2k+2}} - \varepsilon & (p, q) = (i_{2k+1}, i_{2k+2}) \\
    \pi_{i_{2k+3} i_{2k+2}} + \varepsilon & (p, q) = (i_{2k+3}, i_{2k+2}) \\
    \pi_{pq} & \text{otherwise}\\
    \end{cases}
\end{align} or \begin{align}
    \pi''_{pq} = \begin{cases}
    \pi_{i_{2k+1} i_{2k+2}} + \varepsilon & (p, q) = (i_{2k+1}, i_{2k+2}) \\
    \pi_{i_{2k+3} i_{2k+2}} - \varepsilon & (p, q) = (i_{2k+3}, i_{2k+2}) \\
    \pi_{pq} & \text{otherwise}\\
    \end{cases}
\end{align} is no worse than $\boldPi^*$, where $\varepsilon = \min \{\pi_{i_{2k+1} i_{2k+2}} \mid k = 0, \cdots, n-1\} \cup \{\pi_{i_{2k+3} i_{2k+1}} \mid k = 0, \cdots, n-1\}$, and both supports of $\pi'_{pq}$ and $\pi''_{pq}$ do not have this cycle. Therefore, we assume the support of $\boldP^*$ has no cycle without loss of generality, i.e., the graph is a tree. Let $\mathcal{T}$ be the graph induced by the support of $\boldPi^*$ and let the weight of edge $e = (i, j)$ be $w_e = d(w_i, w_j)$. As the shortest path distance between $u$ and $v$ on tree $\mathcal{T}$ is no less than $d(u, v)$ by definition, $W_1(\mu, \nu) \le W_\mathcal{T}(\mu, \nu)$ holds. However, any pair $(i, j)$ on the support of $\boldPi^*$ is directly connected in $\mathcal{T}$. Therefore, \begin{align*}
    W_1(\mu, \nu) &= \min_{\boldPi \in U(\mu,\nu)}\hspace{.3cm} \sum_{i=1}^{n}\sum_{j=1}^{n'} \pi_{ij}  d(\boldx_i, \boldx'_j), \\
    &= \sum_{i=1}^{n}\sum_{j=1}^{n'} \pi^*_{ij}  d(\boldx_i, \boldx'_j) \\
    &= \sum_{i=1}^{n}\sum_{j=1}^{n'} \pi^*_{ij}  d_\mathcal{T}(\boldx_i, \boldx'_j) \\
    &\ge \min_{\boldPi \in U(\mu,\nu)}\hspace{.3cm} \sum_{i=1}^{n}\sum_{j=1}^{n'} \pi_{ij}  d_\mathcal{T}(\boldx_i, \boldx'_j) \\
    &= W_\mathcal{T}(\mu, \nu).
\end{align*} Therefore, $W_1(\mu, \nu) = W_\mathcal{T}(\mu, \nu)$ and $\boldP^*$ is an optimal solution for both $W_1(\mu, \nu)$ and $W_\mathcal{T}(\mu, \nu)$.

\end{document}